\documentclass[preprint,12pt]{elsarticle}

\usepackage[utf8]{inputenc}
\usepackage[T1]{fontenc}
\usepackage{graphicx}
\usepackage{amsmath}
\usepackage{float}
\usepackage{placeins} 
\usepackage{caption}
\usepackage{subcaption}
\usepackage{comment}
\usepackage[colorlinks=true, allcolors=blue]{hyperref}
\usepackage{cleveref}

\begin{document}

\title{Information Geometry of Evolution of Neural Network Parameters While Training}

\author[1,3]{Abhiram Anand Thiruthummal\corref{cor1}}
\ead{thiruthuma@uni.coventry.ac.uk}
\author[1,4]{Eun-jin Kim}
\ead{ejk92122@gmail.com}
\author[2,3]{Sergiy Shelyag}
\ead{sergiy.shelyag@flinders.edu.au}
\cortext[cor1]{Corresponding author}
\affiliation[1]{
organization={Centre for Fluids and Complex Systems, Coventry University},
city={Coventry},
country={United Kingdom}
}
\affiliation[2]{
organization={College of Science and Engineering, Flinders University},
city={Adelaide},
country={Australia}
}
\affiliation[3]{
organization={School of Information Technology, Deakin University},
city={Melbourne},
country={Australia}
}
\affiliation[4]{
organization={Nuclear Research Institute for Future Technology and Policy, Seoul National University},
city={Seoul},
country={Korea}
}

\begin{abstract}
    Artificial neural networks (ANNs) are powerful tools capable of approximating any arbitrary mathematical function, but their interpretability remains limited, rendering them as black box models. To address this issue, numerous methods have been proposed to enhance the explainability and interpretability of ANNs. In this study, we introduce the application of information geometric framework to investigate phase transition-like behavior during the training of ANNs and relate these transitions to overfitting in certain models.
    
    {
    The evolution of ANNs during training is studied by looking at the probability distribution of its parameters. Information geometry utilizing the principles of differential geometry, offers a unique perspective on probability and statistics by considering probability density functions as points on a Riemannian manifold. We create this manifold using a metric based on Fisher information to define a distance and a velocity. By parameterizing this distance and velocity with training steps, we study how the ANN evolves as training progresses. Utilizing standard datasets like MNIST, FMNIST and CIFAR-10, we observe a transition in the motion on the manifold while training the ANN and this transition is identified with over-fitting in the ANN models considered. The information geometric transitions observed is shown to be mathematically similar to the phase transitions in physics. Preliminary results showing finite-size scaling behavior is also provided. This  work contributes to the development of robust tools for improving the explainability and interpretability of ANNs, aiding in our understanding of the variability of the parameters these complex models exhibit during training.}

\end{abstract}


\maketitle

\section{Introduction}
Artificial neural networks (ANNs) are universal function approximators \cite{hornik1989multilayer, kidger2020universal} capable of learning any arbitrary function. Training an ANN involves optimizing a loss function w.r.t. the parameters of the ANN. Even though ANNs can be trained and used for inference on new data, very little is known about how and what ANNs are learning based on its parameters. Therefore ANNs are often considered as black box models \cite{castelvecchi2016can}. Several methods have been previously proposed to make ANNs more explainable and interpretable \cite{linardatos2020explainable}. With AI becoming an integral part of our lives and considering the societal impact of large language models like ChatGPT \cite{abdullah2022chatgpt}, there is a demand for the development of more robust and versatile tools to improve the explainability and interpretability of ANNs. In this work, we introduce the use of information geometric framework as a tool to improve interpretability of the training process for ANNs.

{
Some of the first studies using information geometry in the context of ANNs can be attributed to Shun'ichi Amari who also founded the field of information geometry in its modern form \cite{amari1983foundation}. In \cite{amari1998natural} Amari used information geometry to develop the Natural Gradient algorithm for training neural networks. Even though the algorithm has desired theoretical properties, it is seldom used in practice due to the difficulty in computing the natural gradient term. Amari has also used information geometry to study the Expectation Maximization algorithm for training neural networks \cite{amari1995information}. A more recent approach to developing an information geometry-inspired training algorithm is the Fisher SAM algorithm \cite{kim2022fisher}, which is shown to outperform other similar methods. Information geometry was used to understand the training process of ANNs \cite{sokol2018information}. The framework was used to understand why orthogonal weight initialization of ANNs leads to massive improvement in training speeds. Another recent work in this direction used information geometry to understand the feature selection process in ANNs \cite{xu2022information}. The study also introduced an information geometric measure called the H-score to quantify performance of the features extracted from the data by the ANN. 

In this first of its kind study, we use information geometry to empirically explore the collective behavior of the parameters of a neural network while training. We achieve this by investigating the evolution of probability density function (PDF) of neural network parameters during the training process and studying the transition in its behavior. In order to quantitatively study this transition, we use information geometric measures, such as information length, information velocity and its derivatives. This transition is shown to coincide with overfitting and can in specific cases be used to predict overfitting without referring to the test dataset.

In Sec.~\ref{sec:background} we provide a concise introduction to neural networks and the theory of information geometry including the definitions of information length and information velocity. Sec.~\ref{sec:problem_setup} describes the problem setup, including the datasets and the neural network architecture used. The section also provides the exact numerical prescription used to compute information length and information velocity (Sec. \ref{sec:info_geo_nn}), and introduces the technique of regularized derivatives (Sec. \ref{sec:reg_deriv}) to remove noise from the information velocity estimates and its derivatives, in order to extract useful information. In Sec. \ref{sec:nn_ir_results} we conduct numerical experiments on ANNs and identify a transition in the behavior of information length. This transition is shown to coincide with overfitting in the ANNs. After introducing a quantity to better identify these transitions, for the case of a shallow network trained on MNIST dataset, the effect of different hyperparameters like learning rate (Sec. \ref{sec:lr}), dropout probability (Sec. \ref{sec:dropout}) and strength of noise regularization (Sec. \ref{sec:noise}) is investigated. Similar investigation is also carried out for the Fashion-MNIST dataset in Sec. \ref{sec:fmnist}. In Sec. \ref{sec:resnet}, the transition is shown to exist even in deeper networks. We demonstrate it using a ResNet architecture trained on the CIFAR-10 dataset. In Sec. \ref{sec:il_phase_transition} we draw parallels between the information geometric transitions studied in this work and thermodynamic phase transitions in physics. The information geometric transitions are also shown to exhibit finite-size scaling behavior. Further conclusions and potential for future work are discussed in Sec. \ref{sec:conclusion_nn}.

}

\section{Background}
\label{sec:background}
The training of an ANN is an optimization process. Consider an ANN denoted by $\mathcal{F}(\vec{x}; \vec{\omega})$, where $\vec{x}$ is the input to the ANN and $\vec{\omega}$ is the vector representing the parameters of the ANN. The optimization of the ANN proceeds by first constructing a scalar loss function $L\left(\mathcal{F}\left(\vec{x}; \vec{\omega}\right), \vec{y} \right)$, where $\vec{y}$ is some additional information that might be required to train the ANN. The process of minimizing this loss function is known as training. A simple algorithm used to train ANNs is the gradient descent. Gradient descent works by minimizing the loss function by iteratively updating the parameters, moving in the opposite direction of the gradient,

\begin{equation}
	\vec{\omega}_{t+1} = \vec{\omega}_t - \alpha \nabla_{\vec{\omega}}L\left(\mathcal{F}\left(\vec{x}; \vec{\omega}\right), \vec{y} \right).
\end{equation}
Here $\alpha$ is known as the learning rate (LR), which controls magnitude of the updates. More complex algorithms exist, which adapt the learning rate to improve the optimization process. Some of these algorithms are outlined in Table \ref{tab:hyperparameters}. In this work we study how the distribution of the parameters of ANNs evolves as training progresses, using the framework of information geometry.

Information geometry is the study of probability and statistics using the techniques of differential geometry. In order to do this, we first define a distance metric between probability distributions. Then these probability distributions can be considered as points on a Riemannian manifold. Several different metrics can be defined on a probability space \cite{gibbs2002choosing, majtey2005wootters, diosi1996thermodynamic, gangbo1996geometry}. In this work we use a metric based on Fisher information \cite{frieden2004science} called the Fisher information metric. For a family of PDFs $p(x ;\{\theta\})$ parametrized by $\{\theta\}$, the Fisher information metric $g_{j k}(\{\theta\})$ can be defined as

\begin{equation}
    g_{j k}(\theta):=\int_X \frac{\partial \log p(x ;\{\theta\})}{\partial \theta_j} \frac{\partial \log p(x ;\{\theta\})}{\partial \theta_k} p(x ;\{\theta\}) d x, 
\end{equation}
where $X$ denotes the domain of the PDF. In many physically relevant situations including in this work, we come across time-dependent PDFs $p(x,t)$ which can be considered as PDFs parameterized by variable $t$. In such cases the Fisher information metric simplifies to
\begin{equation}
    g(t)=\int d \mathbf{x} \frac{1}{p(\mathbf{x}, t)}\left[\frac{\partial p(\mathbf{x}, t)}{\partial t}\right]^2.
\end{equation}
Using this metric we can now define information length $\mathcal{L}$ as the total distance travelled on the manifold as the PDF evolves with time:
\begin{equation}
\label{eq:il}
    \mathcal{L}(t):=\int_0^t d t_1 \sqrt{\int d \mathbf{x} \frac{1}{p\left(\mathbf{x}, t_1\right)}\left[\frac{\partial p\left(\mathbf{x}, t_1\right)}{\partial t_1}\right]^2}.
\end{equation}
Information length defined in this way is a dimensionless quantity, which quantifies the number of statistically distinguishable states a system has passed through during its temporal evolution \cite{wootters1981statistical}. Information length has previously been used to study thermodynamics, dynamical systems, phase transitions and self-organization \cite{hollerbach2020time, kim2020time, kim2018investigating, kim2017geometric, heseltine2019comparing, kim2020information, crooks2007measuring, feng2009far}.

We can now define an information velocity $\Gamma := \lim _{d t \rightarrow 0} d \mathcal{L} / d t$ on the manifold (note that this quantity is referred to as information rate in previous works \cite{kim2021causal, kim2021thermodynamic, kim2021information}):
\begin{equation}
\label{eq:gamma}
    \Gamma(t):=\sqrt{\int d \mathbf{x} \frac{1}{p(\mathbf{x}, t)}\left[\frac{\partial p(\mathbf{x}, t)}{\partial t}\right]^2}.
\end{equation}
As for the significance of $\Gamma$, it measures the rate of change of statistically distinguishable states of a time-evolving PDF. Therefore $\Gamma$ has been used to study causality \cite{kim2021causal} and abrupt changes in dynamical systems \cite{guel2021information}. In the context of non-equilibrium thermodynamics, $\Gamma$ is related to the entropy production rate for a Gaussian process \cite{kim2021information}.

In this work we will look at how the probability distribution of parameters of an ANN changes during training using the quantities $\mathcal{L}$ and $\Gamma$. In \cref{sec:lr,sec:dropout,sec:noise,sec:fmnist,sec:resnet} the changes in the values of $\mathcal{L}$ and $\Gamma$ are shown to independently indicate overfitting in a ANN, without referring to the test dataset. In section \ref{sec:il_phase_transition} we show how $\mathcal{L}$ and $\Gamma$ exhibit phase transition like behavior.

\section{Problem setup}
\label{sec:problem_setup}
\subsection{Dataset}

In this study we use three different datasets, MNIST \cite{deng2012mnist}, Fashion-MNIST \cite{xiao2017fashion} and CIFAR-10 \cite{krizhevsky2009learning}. The datasets were chosen due to their popularity as standard datasets for testing ANNs.
The MNIST dataset contains $28 \times 28$ grayscale images of handwritten digits from 0 to 9 with their corresponding labels. The dataset contains 60,000 training samples and 10,000 test samples. Before training, both the training and testing data were normalized by first subtracting 0.1307 and then dividing by 0.3081, the approximate mean and standard deviation, respectively, of the training dataset.

The Fashion-MNIST dataset also contains $28 \times 28$ grayscale images but of 10 different clothing categories and their corresponding labels. The dataset contains 60,000 training samples and 10,000 test samples. Similar to the case of MNIST dataset, the dataset was normalized by first subtracting the mean 0.2860 and then dividing by the  standard deviation 0.3530.

{
The CIFAR-10 dataset is made of 60,000 $32 \times 32$ color images of 10 mutually exclusive classes. Among the 60,000, 10,000 samples are for testing. The dataset is normalized by subtracting (0.4914, 0.4822, 0.4465) and then dividing by (0.2023, 0.1994, 0.2010). Note that we have a 3-tuple here since the images are 3D.}

In Sec. \ref{sec:nn_ir_results} we will primarily focus on the MNIST dataset and reproduce the same results for the Fashion-MNIST dataset in Sec. \ref{sec:fmnist}. {CIFAR-10 is used in Sec. \ref{sec:resnet} to train a ResNet architecture. }

\subsection{Neural Network Training}
The ANN used in this work is a simple fully connected network with Leaky ReLU activation function. The specific choice of activation function is not important as discussed in Sec. \ref{sec:nn_ir_results}. The choice of 4096 neurons in the hidden layer was deliberate. Since the input vector has a length of 784, there will be $784 \times 4096 = 3211264$ parameters between the input layer and the hidden layer. The large number of parameters will produce less noisy PDFs and hence smoother information length estimates. The neural network outputs vectors of length 10 containing the logarithm of the prediction probability corresponding to each of the 10 categories in MNIST or Fashion-MNIST. The ANN is trained by minimizing the negative log-likelihood loss function.
\begin{figure}[h]
  \includegraphics[width=\linewidth]{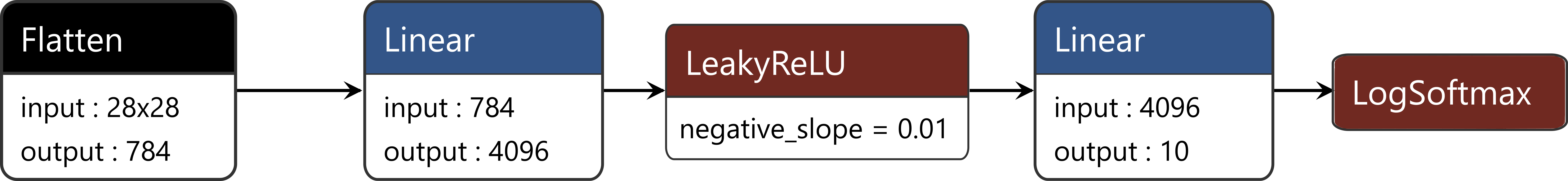}
  \caption{Neural architecture used in this work for MNIST and Fashion-MNIST classification task.}
  \label{fig:nn}
\end{figure}

\subsection{Information Geometry}
\label{sec:info_geo_nn}
While training an ANN, each training step consists of a forward propagation, a backward propagation and an update of parameters. If we consider the distribution of values of these parameters, each training step changes this distribution. We can quantify this change using information length and information velocity. Using this method we can study the distribution of the entire set of parameters of an ANN or a subset of interest.

The form of information length and information velocity given by Eq. \ref{eq:il} and Eq. \ref{eq:gamma}, respectively, is not numerically stable due to the presence of the term $1/p(x,t)$ in the integrand, which blows up when $p(x,t) \to 0$. We can derive a numerically stable form of Eq. \ref{eq:il} by using the definition $q(x,t) := \sqrt{p(x,t)}$. $\Gamma$ and $\mathcal{L}$ can then be written using $q(x,t)$ as
\begin{align}
    \label{eq:il_q}
    \mathcal{L}(t) & = 2 \int_{0}^{t} dt_1 \sqrt{\int d \mathbf{x}\left[\frac{\partial q(\mathbf{x}, t_1)}{\partial t_1}\right]^2},\\
    \label{eq:gamma_q}
    \Gamma(t) & = 2 \sqrt{\int d \mathbf{x}\left[\frac{\partial q(\mathbf{x}, t)}{\partial t}\right]^2}.
\end{align}

\begin{figure}[H]
  \begin{subfigure}[b]{0.49\textwidth}
    \includegraphics[width=\textwidth]{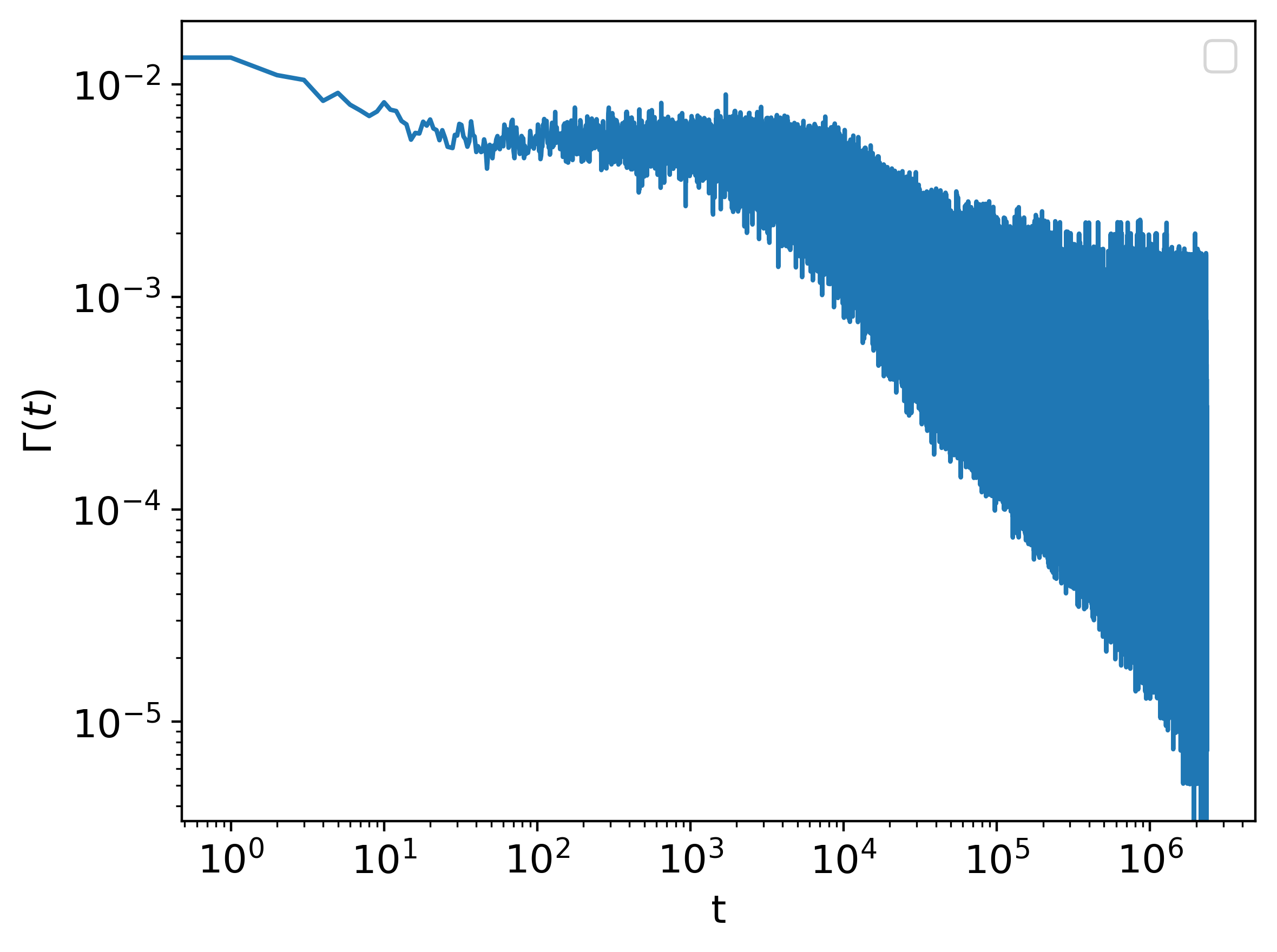}
  \end{subfigure}
  \hfill
  \begin{subfigure}[b]{0.49\textwidth}
    \includegraphics[width=\textwidth]{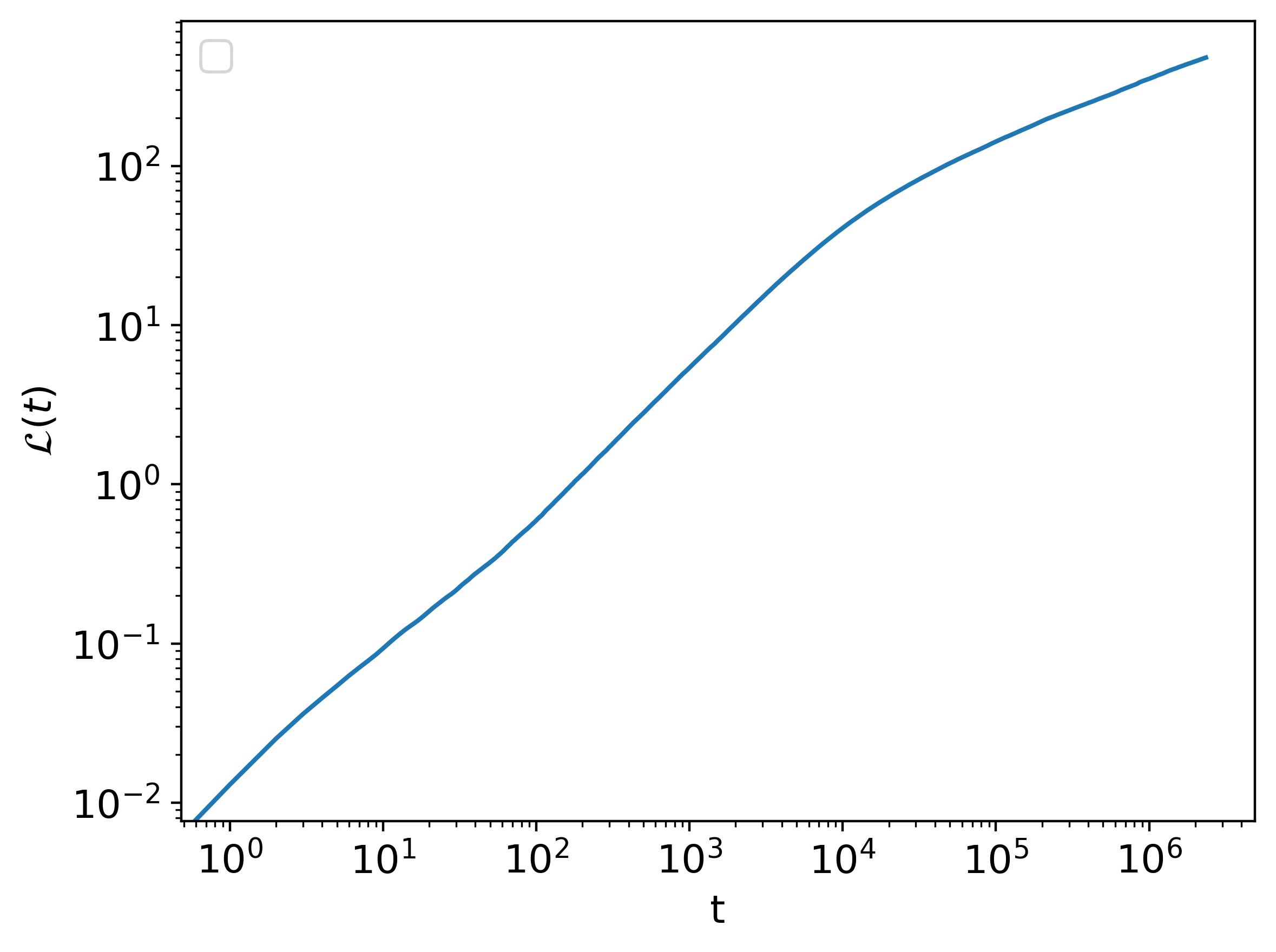}
  \end{subfigure}
  \caption{An example of \textbf{(left)} information velocity  and \textbf{(right)} information length estimates for the network described in Fig. \ref{fig:nn} trained on MNIST data using SGD optimizer with learning rate 0.045.}
  \label{fig:il_gamma_example}
\end{figure}

In order to calculate the information length we first need to estimate the PDF of the parameters. This estimate is done using a histogram. Computing the histogram is an inexpensive method of approximating the probability distribution of a large number of samples, with $O(n)$ time complexity, where $n$ is the number of samples. It can also be efficiently parallelized, making it suitable for GPU computations \cite{shams2007efficient}. Kernel density estimation (KDE) \cite{wkeglarczyk2018kernel} is another approach which is computationally more expensive \cite{raykar2010fast} but produces smoother PDFs with less bias. In this work we use histograms since we are estimating PDFs using more than 1 million samples and KDE will be significantly slower with marginal improvement in accuracy. The number of bins in the histogram are chosen by an empirical formula $2.59 \sqrt[3]{n}$ which is similar to Rice's rule \cite{terrell1985oversmoothed}.
\\
Once the PDF is estimated the information velocity $\Gamma$ is computed using Eq. \ref{eq:gamma_q} by approximating the derivative as a finite difference and integral using trapezoidal rule. Note that $dt$ between consecutive training steps are chosen to be 1. $\Gamma(t)$ is then integrated using trapezoidal rule to obtain information length $\mathcal{L}(t)$. Fig. \ref{fig:il_gamma_example} shows an example of information velocity and information length estimates.

\subsection{Regularized derivative}
\label{sec:reg_deriv}
In Fig. \ref{fig:il_gamma_example} \textbf{(left)}, significant fluctuations can be seen in the value of $\Gamma$. These fluctuations can drown out essential features of interest. Also in Sec. \ref{sec:exp_res} we will look at various 1\textsuperscript{st} and 2\textsuperscript{nd} order derivatives of functions of $\mathcal{L}$. Note that the 1\textsuperscript{st} derivative of $\mathcal{L}$ is $\Gamma$. Then the second derivative will be the derivative of the already noisy $\Gamma$. This will result in even more noise in the derived value. Therefore we need a better approach to computing derivatives which can discern the essential features while ignoring the noise. To solve this, based on \cite{chartrand2011numerical}, we use an optimization-based technique to compute the derivatives, which can then be regularized to reduce the noise.

Assume $u(t)$ is the derivative of $\mathcal{L}(t)$. Then by definition, $\mathcal{L}(t) = \mathcal{L}(0) + \int_0^t u(t_1) dt_1$. Therefore $u(t)$ will be the solution that minimizes the following loss function:

\begin{equation}
    \operatorname{Loss} \left[ u(t) \right] = \frac{1}{2} \int_{0}^{t}  \left|\mathcal{L}(0) + \int_{0}^{t_1} u\left(t_2\right) dt_2 -\mathcal{L}(t_1)\right|^{2} dt_1.
\end{equation}
A regularization term is now added to this loss function to enforce continuity and reduce the fluctuations in $u(t)$:
\begin{equation}
    \operatorname{R-Loss} \left[ u(t) \right] = \lambda \int_0^t \left| u'(t_1) \right|dt_1 + \frac{1}{2} \int_{0}^{t}  \left|\mathcal{L}(0) + \int_{0}^{t_1} u\left(t_2\right) dt_2 -\mathcal{L}(t_1)\right|^{2} dt_1.
\end{equation}
Here $\lambda$ is a parameter which controls the regularization strength. The higher the $\lambda$ the smoother the derivative will be. In this study $\lambda$ is chosen by visually inspecting the plots and choosing a value which can evince the features of interest. During implementation, the integrals are converted into sums using the trapezoidal rule and the derivative is approximated by a finite difference. In the entirety of Sec. \ref{sec:exp_res} we use $\lambda = 0.3$, and to make the computations faster the data is sub sampled to 500 data points equally spaced in the domain of the variable with respect to which the derivative is taken.

\begin{figure}[H]
\centering
  \includegraphics[width=0.6\linewidth]{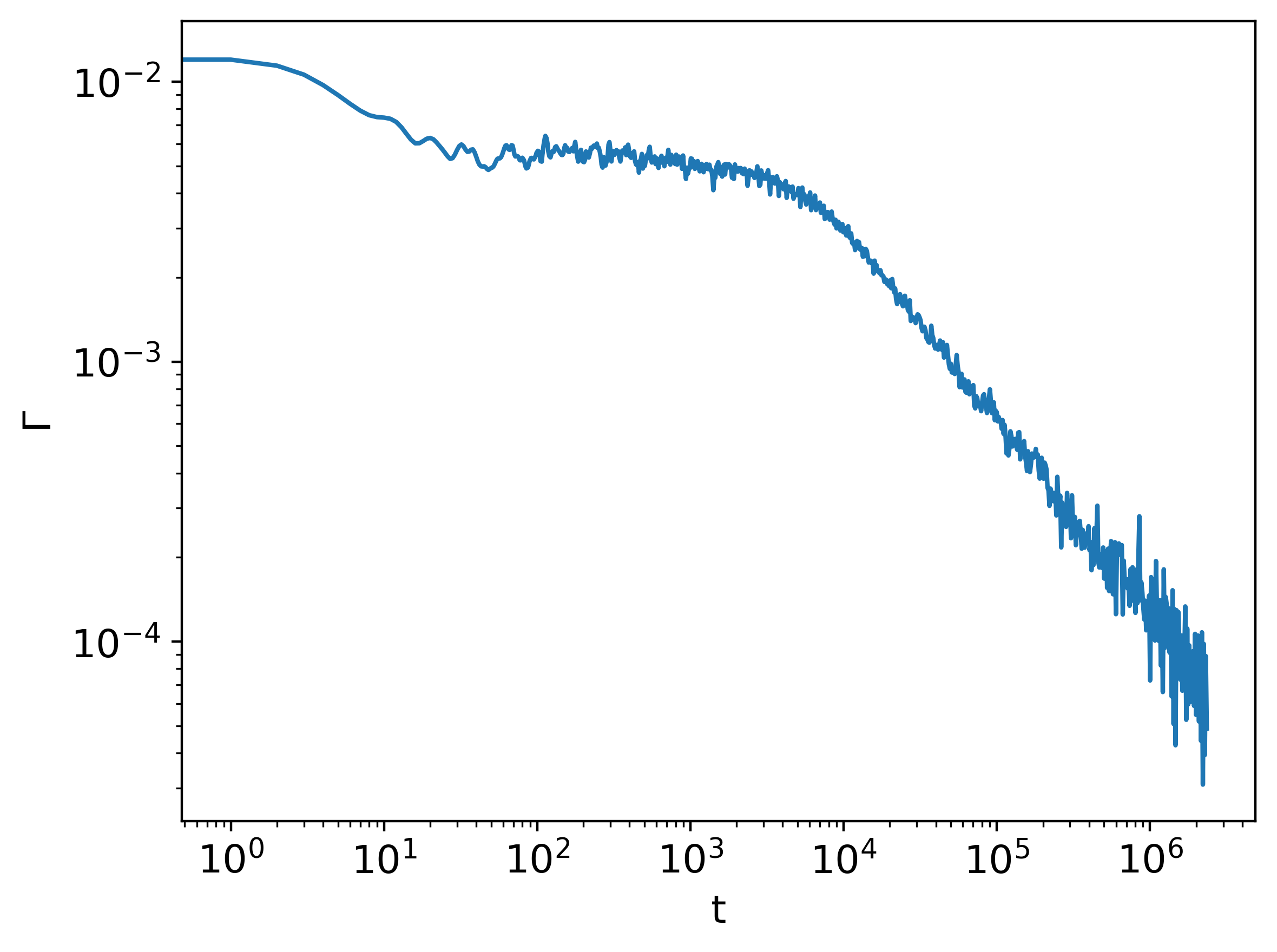}
  \caption{$\Gamma$ calculated from $\mathcal{L}$ shown in Fig. \ref{fig:il_gamma_example} using regularized derivative. Here $\lambda = 100$.}
  \label{fig:sgd_reg_deriv}
\end{figure}

Fig. \ref{fig:sgd_reg_deriv} shows the regularized derivative of information length shown in Fig. \ref{fig:il_gamma_example} \textbf{(right)}. For this plot $5000$ equally spaced data points were chosen from the $\log t$ axis to compute the derivative. Note that compared to \ref{fig:il_gamma_example} \textbf{(left)} $\Gamma$ has significantly reduced noise in Fig. \ref{fig:sgd_reg_deriv}.

\section{Experiments \& Results}
\label{sec:nn_ir_results}
\FloatBarrier
All the numerical experiments conducted in this section have been implemented using PyTorch \cite{pytorch} library for Python. All the neural network optimizers used in this section are from the standard implementation in PyTorch.
\label{sec:exp_res}
\begin{figure}[h]
\centering
  \includegraphics[width=0.8\linewidth]{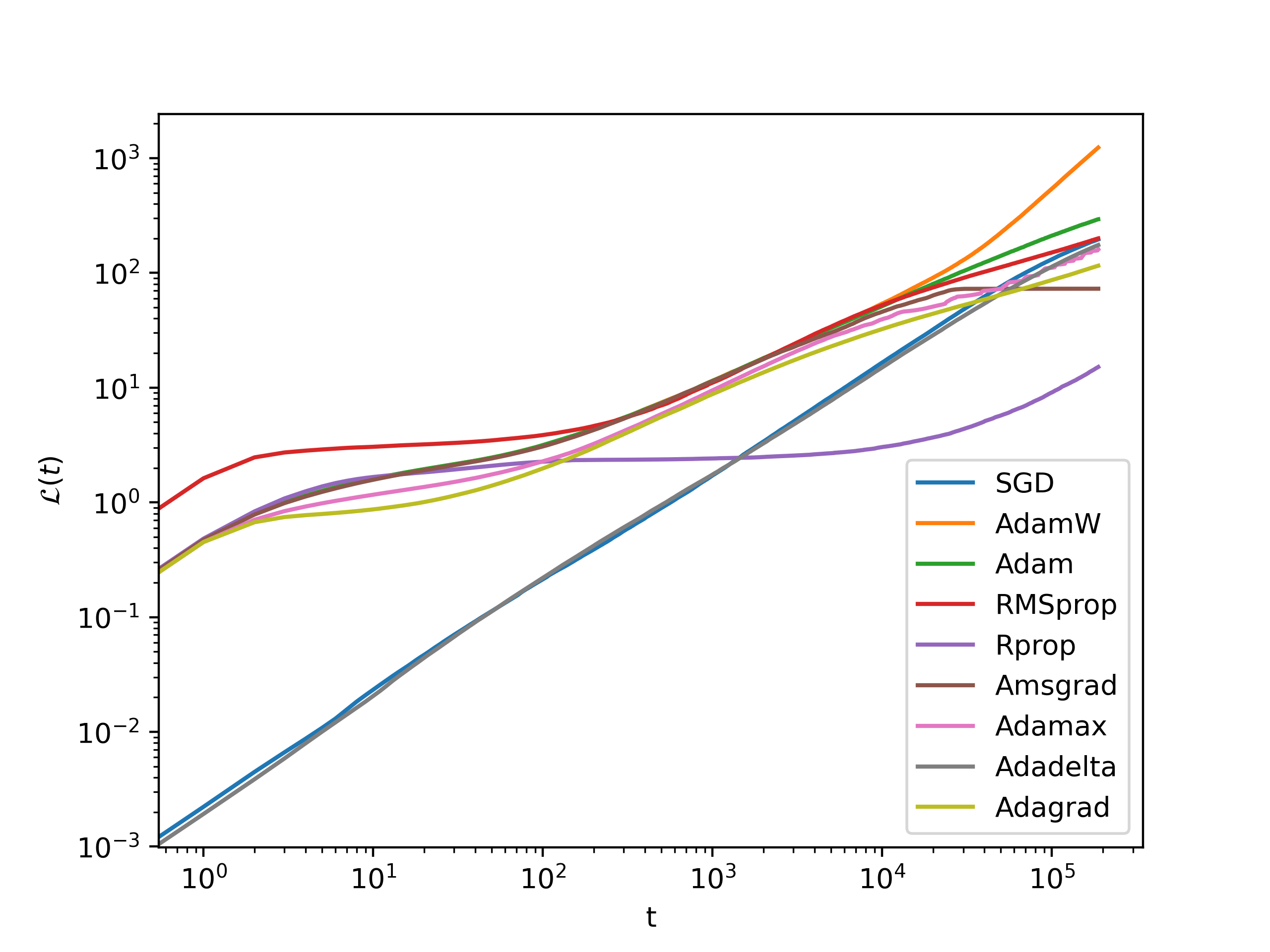}
  \caption{Information Length of different optimizers for the MNIST dataset. The activation function used was Leaky ReLU and learning rate was 0.013.}
  \label{fig:optimizers}
\end{figure}

\begin{figure}[h]
  \begin{subfigure}[b]{0.49\textwidth}
    \includegraphics[width=\textwidth]{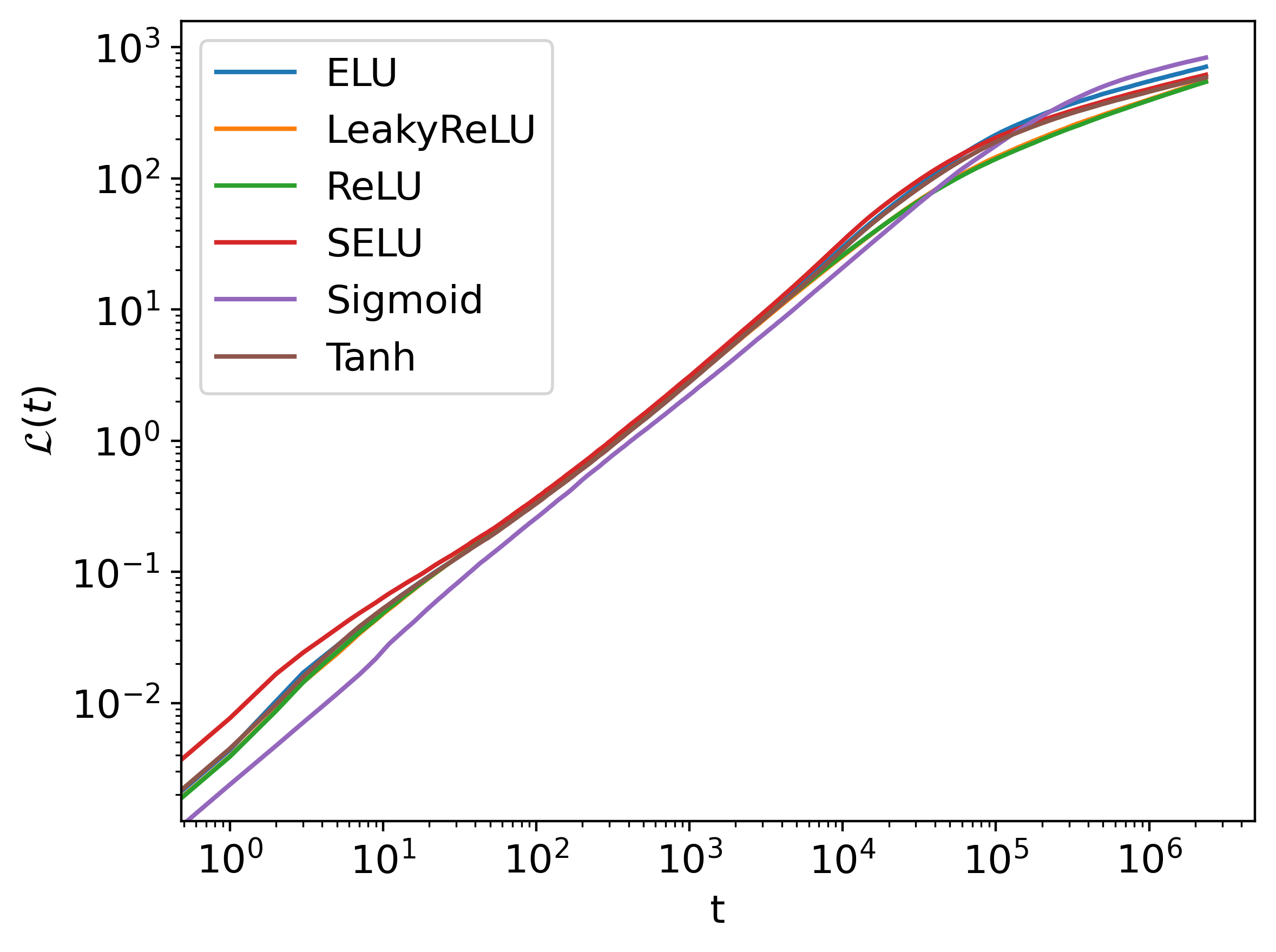}
  \end{subfigure}
  \hfill
  \begin{subfigure}[b]{0.49\textwidth}
    \includegraphics[width=\textwidth]{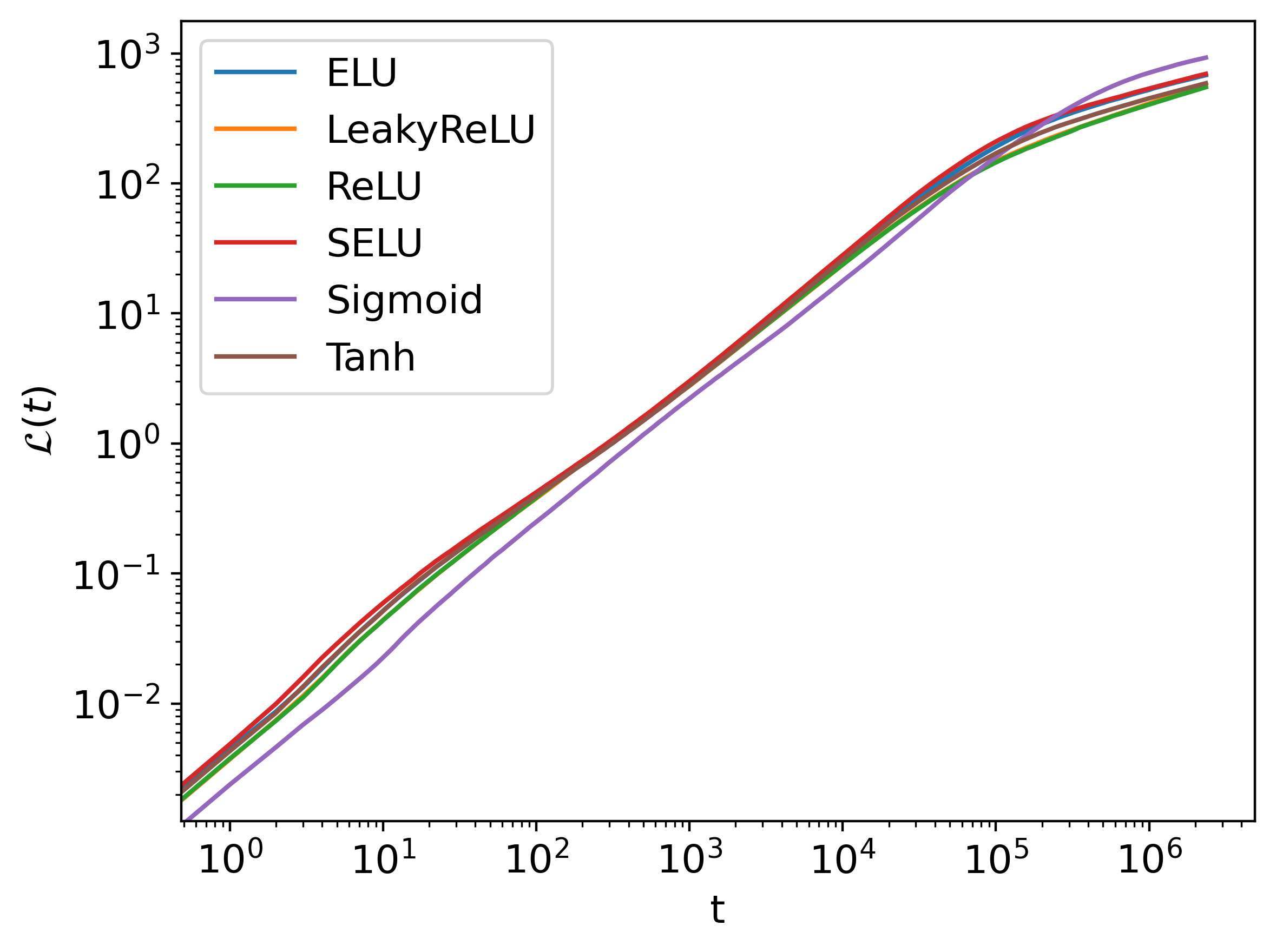}
  \end{subfigure}
  \caption{Information length for  \textbf{(left)} SGD and \textbf{(right)} Adadelta optimizers with different activation functions trained on MNIST dataset.}
  \label{fig:mnist_activation_il}
\end{figure}

Fig. \ref{fig:optimizers} shows the evolution of information length for different optimizers with hyperparameters given in Table. \ref{tab:hyperparameters}. Different optimizers exhibit different behavior. Among the various optimizers SGD and Adadelta exhibit the simplest behavior with $\mathcal{L} \sim t$ for the initial values of $t$ with slope decreasing slightly towards the end of the training curve. Adaptive learning rate algorithms which adjust its learning rate as training progresses show more complex behavior, except for the case of Adadelta. This trend is irrespective of the specific activation function used in the ANN, as shown in Fig. \ref{fig:mnist_activation_il}. A theoretical exposition of these trends is beyond the scope of this study. In the rest of this section we explore in detail how the optimizers with the simplest behavior—SGD and Adadelta—behave when different hyperparameters are changed. Note that hyperparameters are the parameters of a neural network or an optimizer which are not learned during training.

\begin{table}[h]
  \centering
    \begin{tabular}{|l|l|}
    \hline
    \textbf{Optimizer} & \textbf{Hyperparameters} \\
    \hline
    Adadelta \cite{zeiler2012adadelta} & $\text{lr} = 0.005,\, \rho = 0.9,\, \epsilon =  10^{-6}$ \\
    \hline
    Adagrad \cite{lydia2019adagrad} & $\text{lr} = 0.005,\, \epsilon =  10^{-10}$ \\
    \hline
    Adam \cite{kingma2014adam}  & $\text{lr} = 0.005,\, \beta_1 = 0.9,\, \beta_2 = 0.999,\, \epsilon =  10^{-8}$ \\
    \hline
    Adamax \cite{kingma2014adam} & $\text{lr} = 0.005,\, \beta_1 = 0.9,\, \beta_2 = 0.999,\, \epsilon =  10^{-8}$ \\
    \hline
    AdamW \cite{loshchilov2018fixing} & $\text{lr} = 0.005,\, \beta_1 = 0.9,\, \beta_2 = 0.999,\, \epsilon =  10^{-8},\, w=0.01$ \\
    \hline
    Amsgrad \cite{reddi2019convergence} & $\text{lr} = 0.005,\, \beta_1 = 0.9,\, \beta_2 = 0.999,\, \epsilon =  10^{-8}$ \\
    \hline
    RMSprop \cite{hinton2012neural} & $\text{lr} = 0.005,\, \alpha = 0.99,\, \epsilon =  10^{-8}$ \\
    \hline
    Rprop \cite{riedmiller1993direct} & $\text{lr} = 0.005,\, \eta^{+} = 0.5,\, \eta^{-}=1.2,\, \Delta_{\max } = 50,\, \Delta_{\min } = 10^{-6}$ \\
    \hline
    SGD   & $\text{lr} = 0.005$ \\
    \hline
    \end{tabular}%
    \caption{The hyperparameters for different optimizers in Fig. \ref{fig:optimizers}} 
  \label{tab:hyperparameters}%
\end{table}%

\subsection{Learning Rate}
\label{sec:lr}
\begin{figure}[h]
  \begin{subfigure}[b]{0.49\textwidth}
    \includegraphics[width=\textwidth]{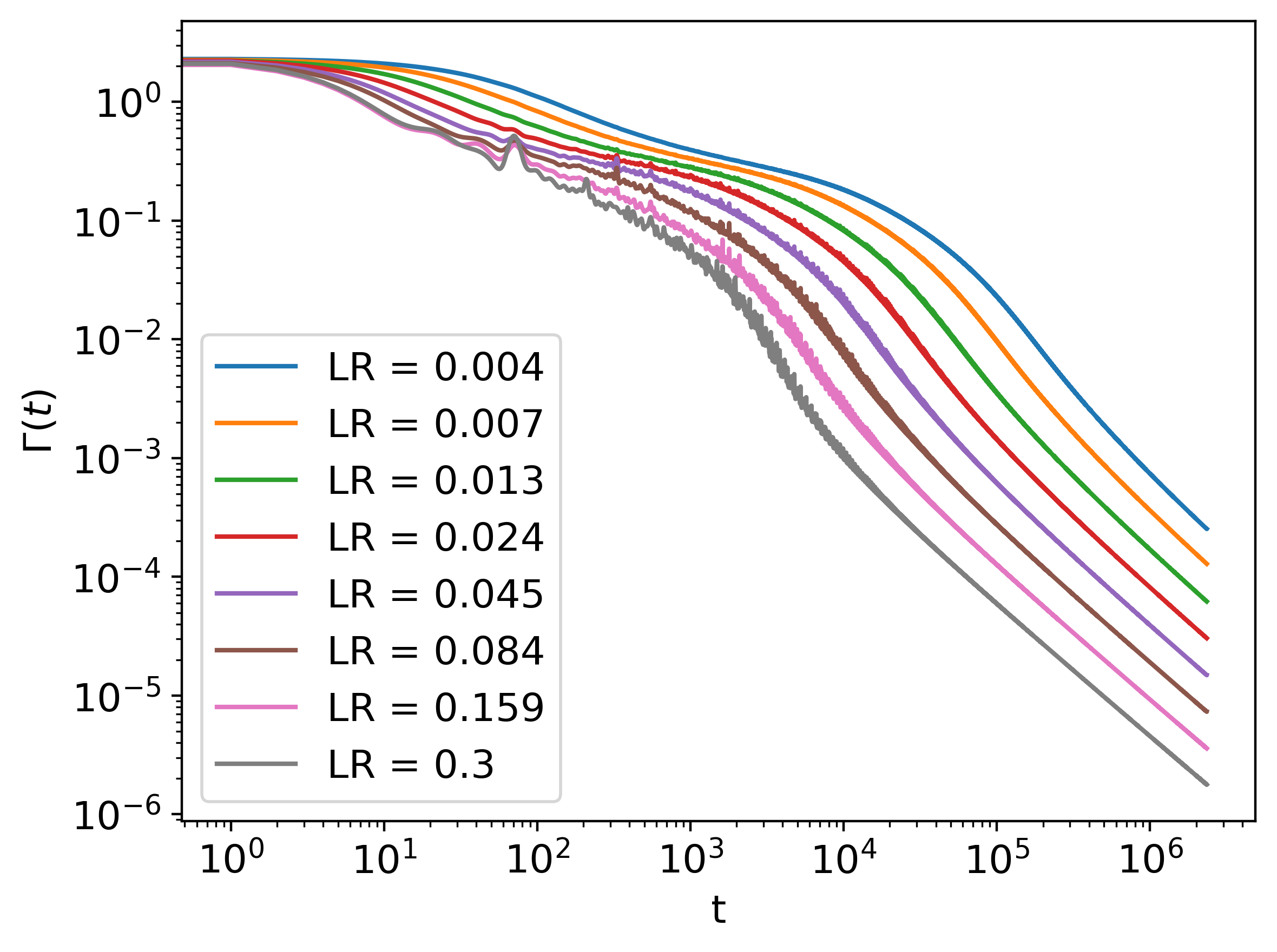}
  \end{subfigure}
  \hfill
  \begin{subfigure}[b]{0.49\textwidth}
    \includegraphics[width=\textwidth]{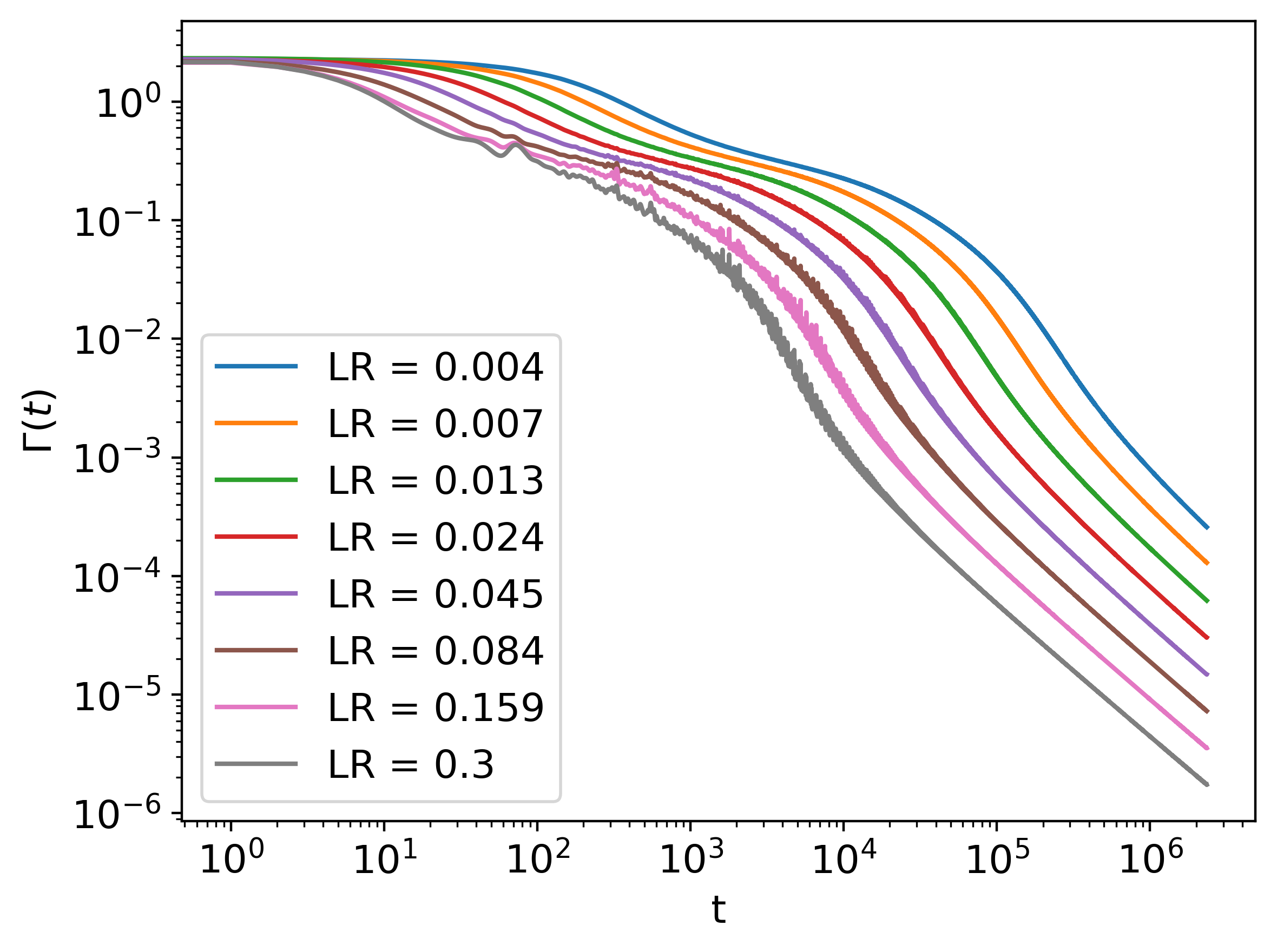}
  \end{subfigure}
  \caption{Loss on training data for \textbf{(left)} SGD and \textbf{(right)} Adadelta optimizers with different learning rates, trained on MNIST dataset.}
  \label{fig:mnist_lr_train_loss}
\end{figure}

\begin{figure}[h]
  \begin{subfigure}[b]{0.49\textwidth}
    \includegraphics[width=\textwidth]{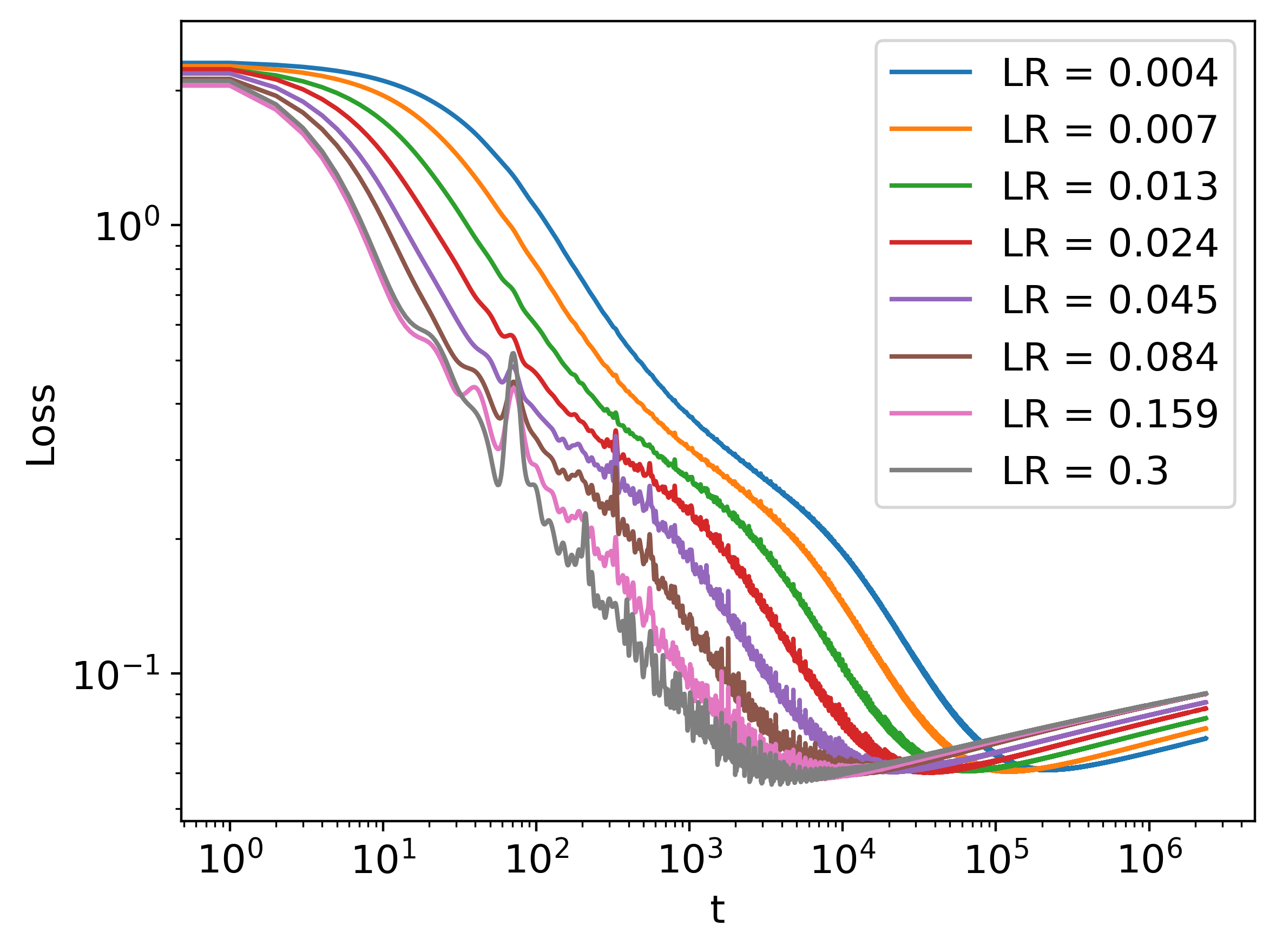}
  \end{subfigure}
  \hfill
  \begin{subfigure}[b]{0.49\textwidth}
    \includegraphics[width=\textwidth]{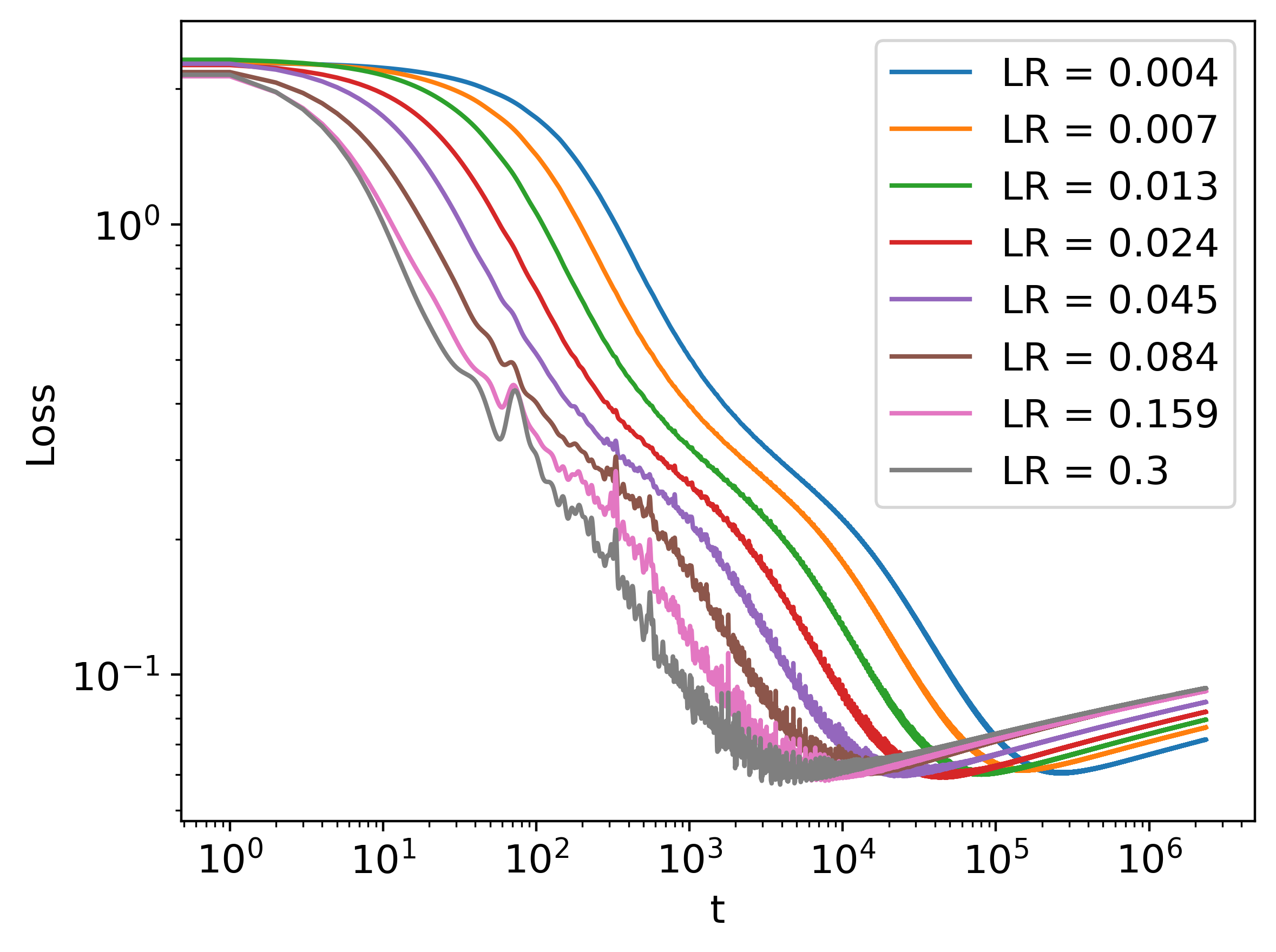}
  \end{subfigure}
  \caption{Loss on test data for \textbf{(left)} SGD and \textbf{(right)} Adadelta optimizers with different learning rates, trained on MNIST dataset.}
  \label{fig:mnist_lr_test_loss}
\end{figure}

Figs. \ref{fig:mnist_lr_train_loss} and \ref{fig:mnist_lr_test_loss} show the value of loss function on the training data and testing data respectively for MNIST dataset. The loss function keeps decreasing as the training progresses for the training data, while for the testing data it first decreases, reaches a minimum, and then increases. This is an example of overfitting: the model fails to generalize well to data it has not been trained on. The minimum is where the neural network has the highest capacity for generalization. The same trend can be seen with other datasets including Fashion-MNIST as shown in Fig. \ref{fig:fmnist_test_loss}.

\begin{figure}[h]
  \begin{subfigure}[b]{0.49\textwidth}
    \includegraphics[width=\textwidth]{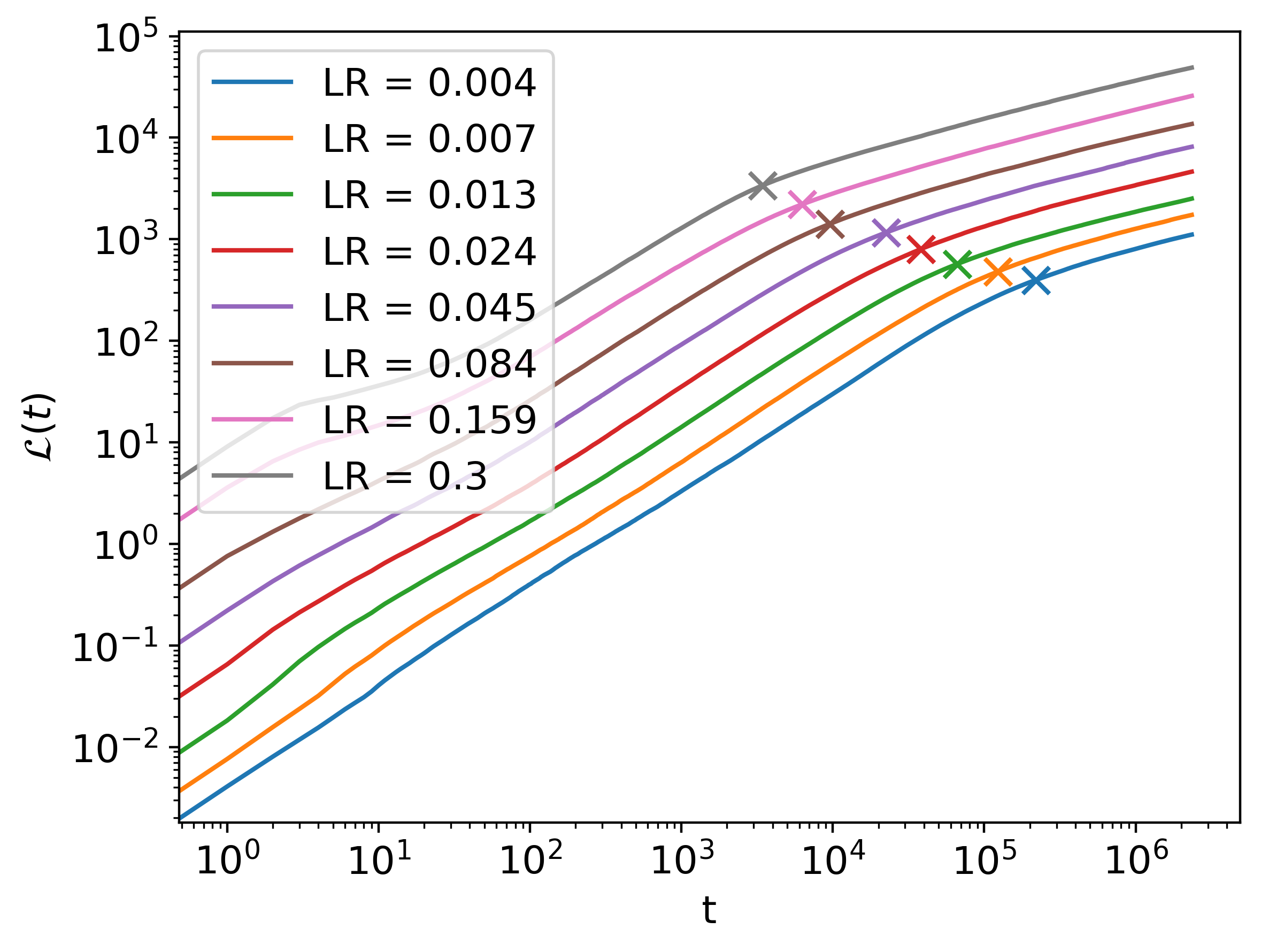}
  \end{subfigure}
  \hfill
  \begin{subfigure}[b]{0.49\textwidth}
    \includegraphics[width=\textwidth]{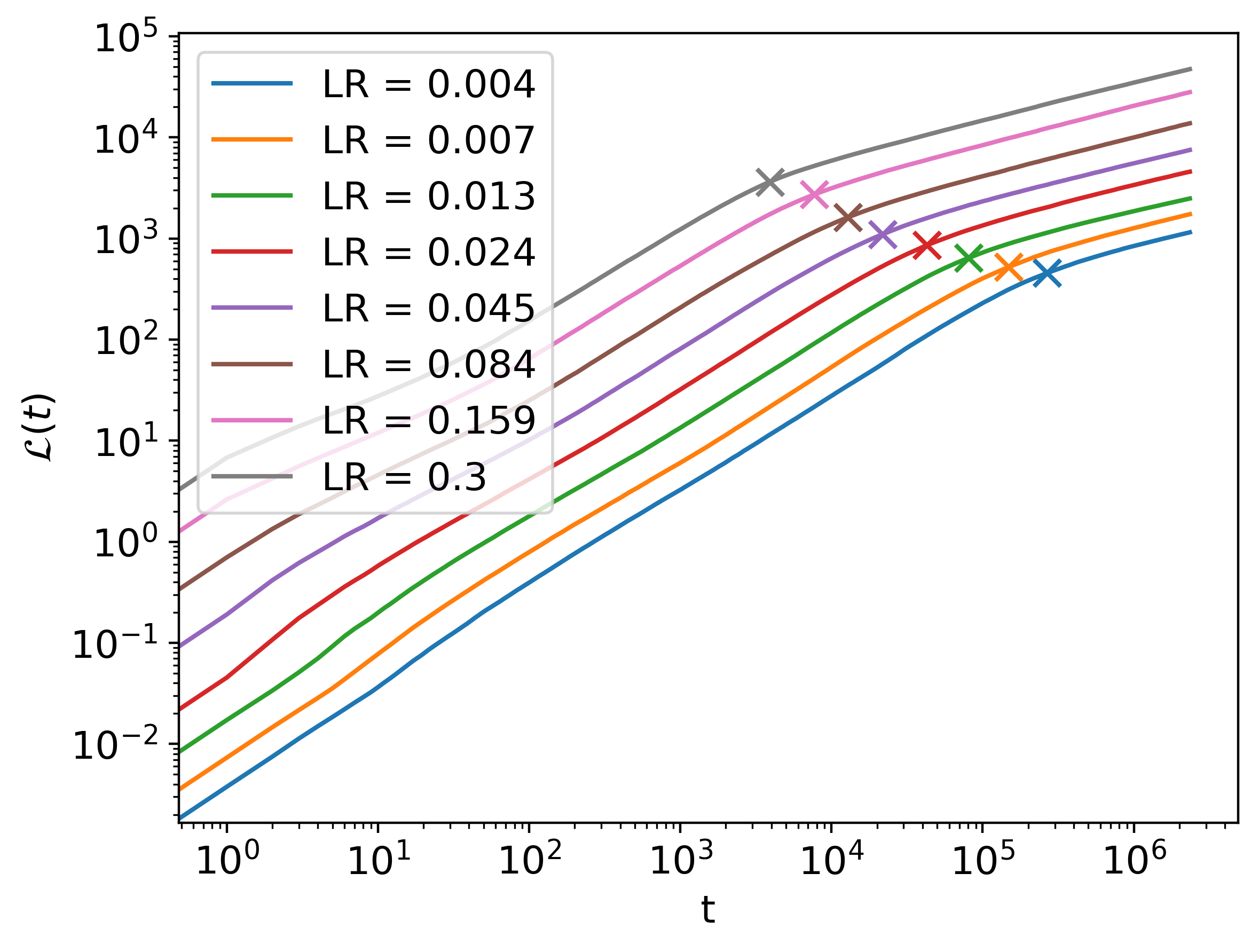}
  \end{subfigure}
  \caption{Information length for SGD \textbf{(left)} and Adadelta \textbf{(right)} optimizers with different learning rates on MNIST dataset. The symbol x represents the minimum of the test loss. Individual curves have been shifted along the y-axis to improve readability. See Fig. \ref{fig:mnist_lr_il_full} for the unmodified plot without the added y-axis shifts.}
  \label{fig:mnist_lr_il}
\end{figure}

Fig. \ref{fig:mnist_lr_il} shows the information length calculated during training corresponding to Fig. \ref{fig:mnist_lr_train_loss} and Fig. \ref{fig:mnist_lr_test_loss}. The training step at which the test loss is minimal is marked on the plot with the symbol 'x'. On a closer inspection, it can be seen that irrespective of the learning rate, 'x' marks the approximate point at which the slope of the information length curve changes. To better quantify the training step at which the slope changes, we compute the 2\textsuperscript{nd} derivative $d^2(\log \mathcal{L}) / d(\log t)^2$. In order to calculate this quantity we first compute the 1\textsuperscript{st} derivative $d(\log \mathcal{L}) / d(\log t)$ using the method outlined in Sec. \ref{sec:reg_deriv} and perform another derivative operation to obtain $d^2(\log \mathcal{L}) / d(\log t)^2$.

\begin{figure}[h]
  \begin{subfigure}[b]{0.49\textwidth}
    \includegraphics[width=\textwidth]{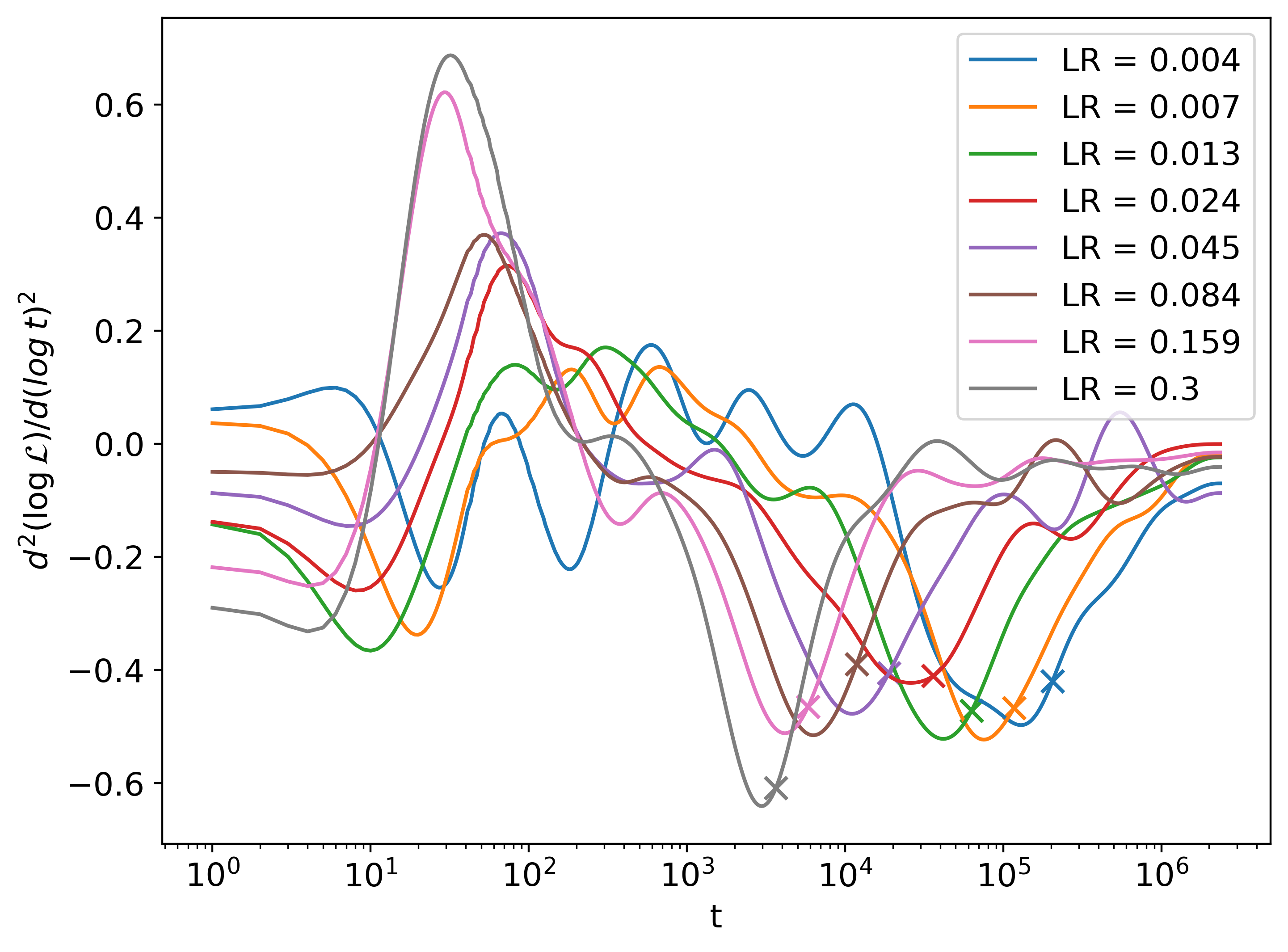}
  \end{subfigure}
  \hfill
  \begin{subfigure}[b]{0.49\textwidth}
    \includegraphics[width=\textwidth]{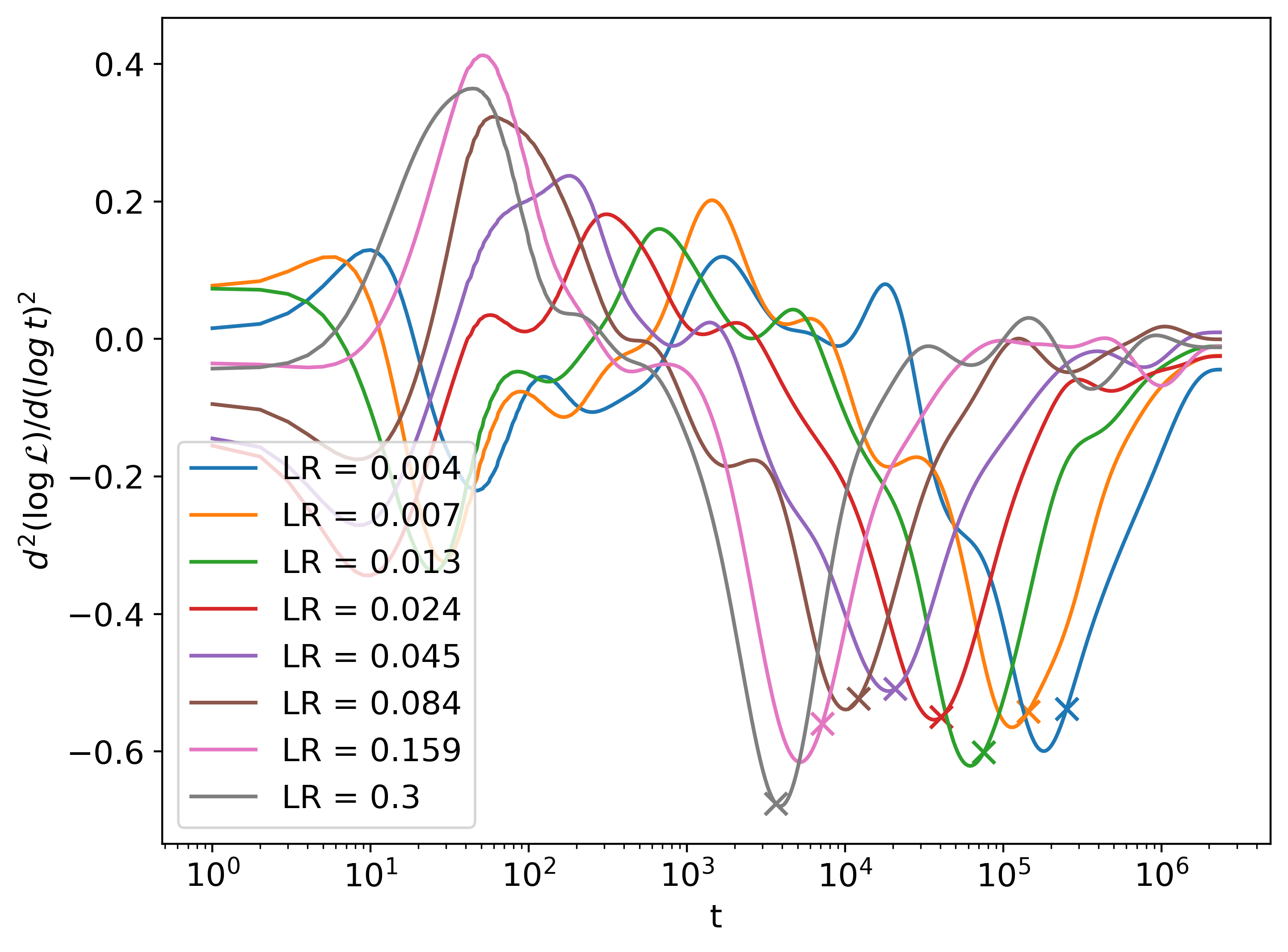}
  \end{subfigure}
  \caption{2\textsuperscript{nd}  derivative of $\log {\mathcal{L}}$ w.r.t $\log {t}$ for \textbf{(left)} SGD  and \textbf{(right)} Adadelta with different learning rates on MNIST dataset. The symbol x represents the minimum of the test loss.}
  \label{fig:lr_deriv}
\end{figure}

Fig. \ref{fig:lr_deriv} shows that for SGD and Adadelta the minima of $d^2(\log \mathcal{L}) / d(\log t)^2$ correspond to overfitting. It should be noted that the information length was computed solely based on the values of the neural network parameters, without any reference to the test dataset. Therefore the minima of $d^2(\log \mathcal{L}) / d(\log t)^2$ act as an independent indicator of overfitting.

Similar trends are also seen with Fashion-MNIST dataset in Sec. \ref{sec:fmnist}.

\subsection{Dropout Regularization}
\label{sec:dropout}
\FloatBarrier
\begin{figure}[h]
    \begin{subfigure}[b]{0.49\textwidth}
    \includegraphics[width=\textwidth]{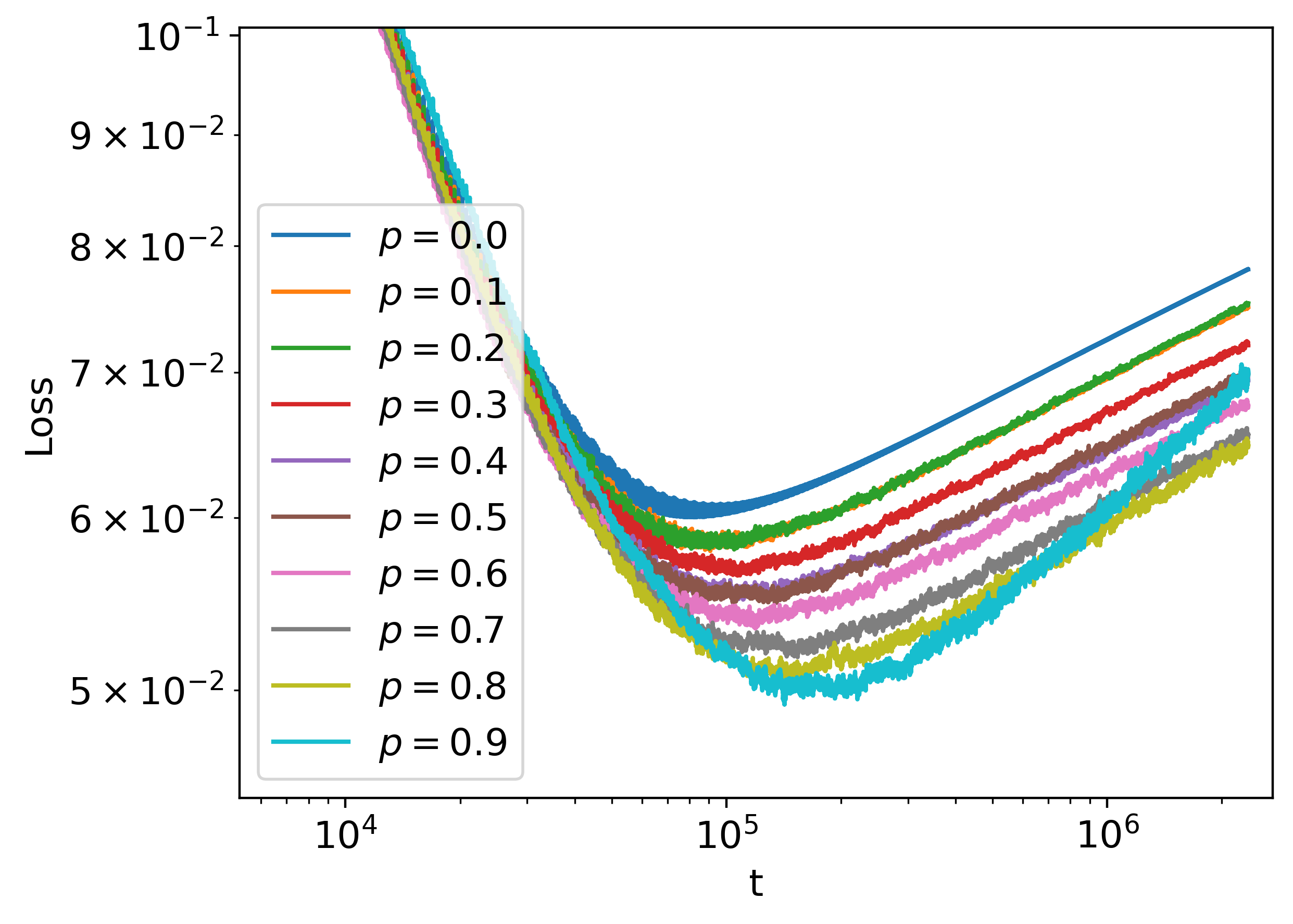}
  \end{subfigure}
  \hfill
  \begin{subfigure}[b]{0.49\textwidth}
    \includegraphics[width=\textwidth]{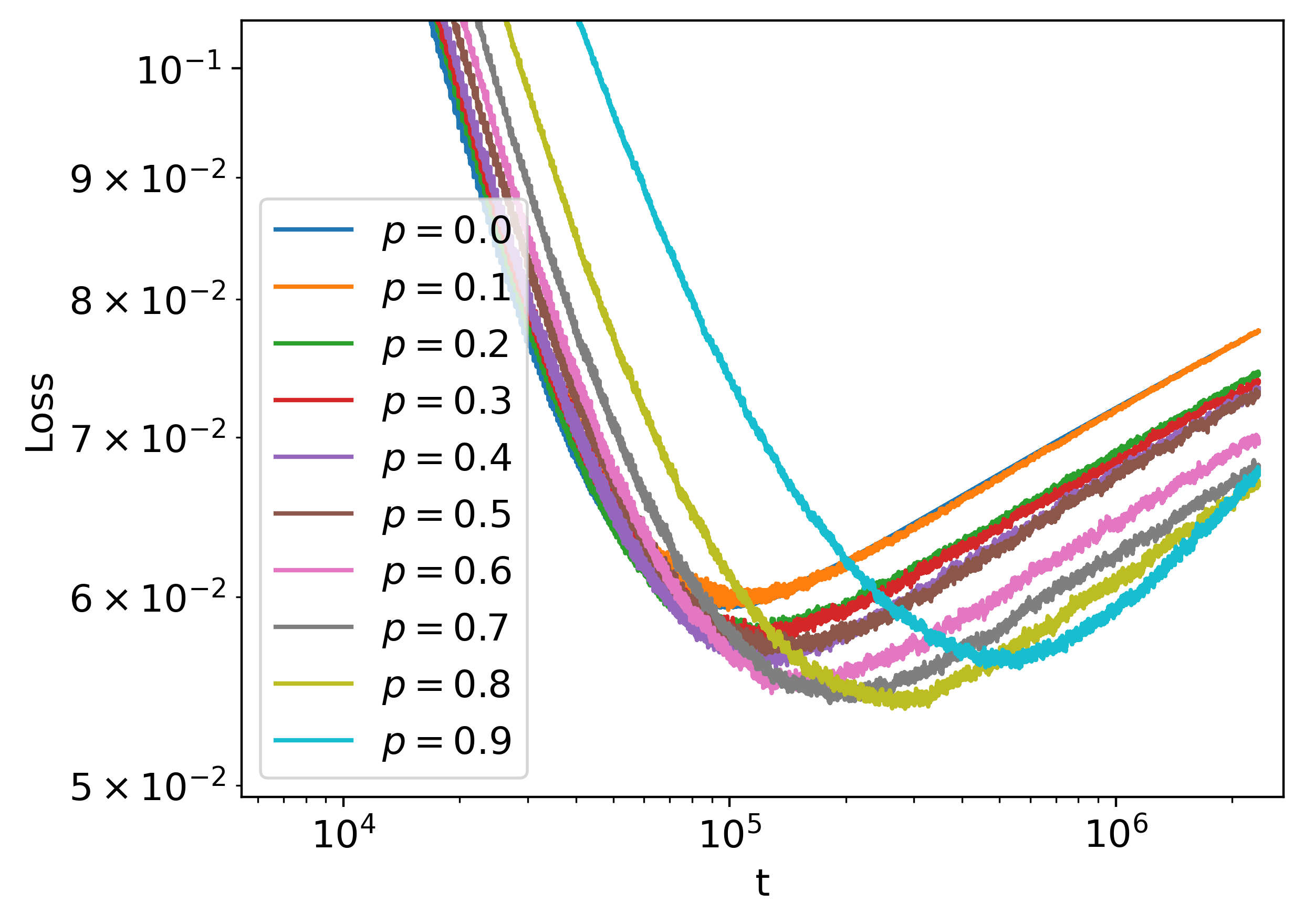}
  \end{subfigure}   
  \caption{Loss on test data for \textbf{(left)} SGD  and \textbf{(right)} Adadelta optimizers and different dropout probabilities on MNIST dataset. The learning rate is 0.013}
  \label{fig:sgd_dropout_loss}
\end{figure}
Dropout regularization was introduced \cite{hinton2012improving} to reduce overfitting in large ANNs. By randomly turning off or dropping out hidden neurons with a probability $p$, the ANN is forced to learn more robust features. This can be seen in Fig.~\ref{fig:sgd_dropout_loss} where the higher the dropout probability, the smaller the minimum test loss, or the model is better at generalizing. Note that the behavior of $p=0.9$ in Fig. \ref{fig:sgd_dropout_loss} is different from other values of $p$. This is because too many neurons are being turned off during the training process, leading to reduced learning capacity. For $p=0.9$, the sharper increase in test loss for SGD and higher value of minimum test loss compared to $p=0.8$ for Adadelta are due to the reduction in the learning capacity, rather than the increase in overfitting.

\begin{figure}[h]
  \begin{subfigure}[b]{0.49\textwidth}
    \includegraphics[width=\textwidth]{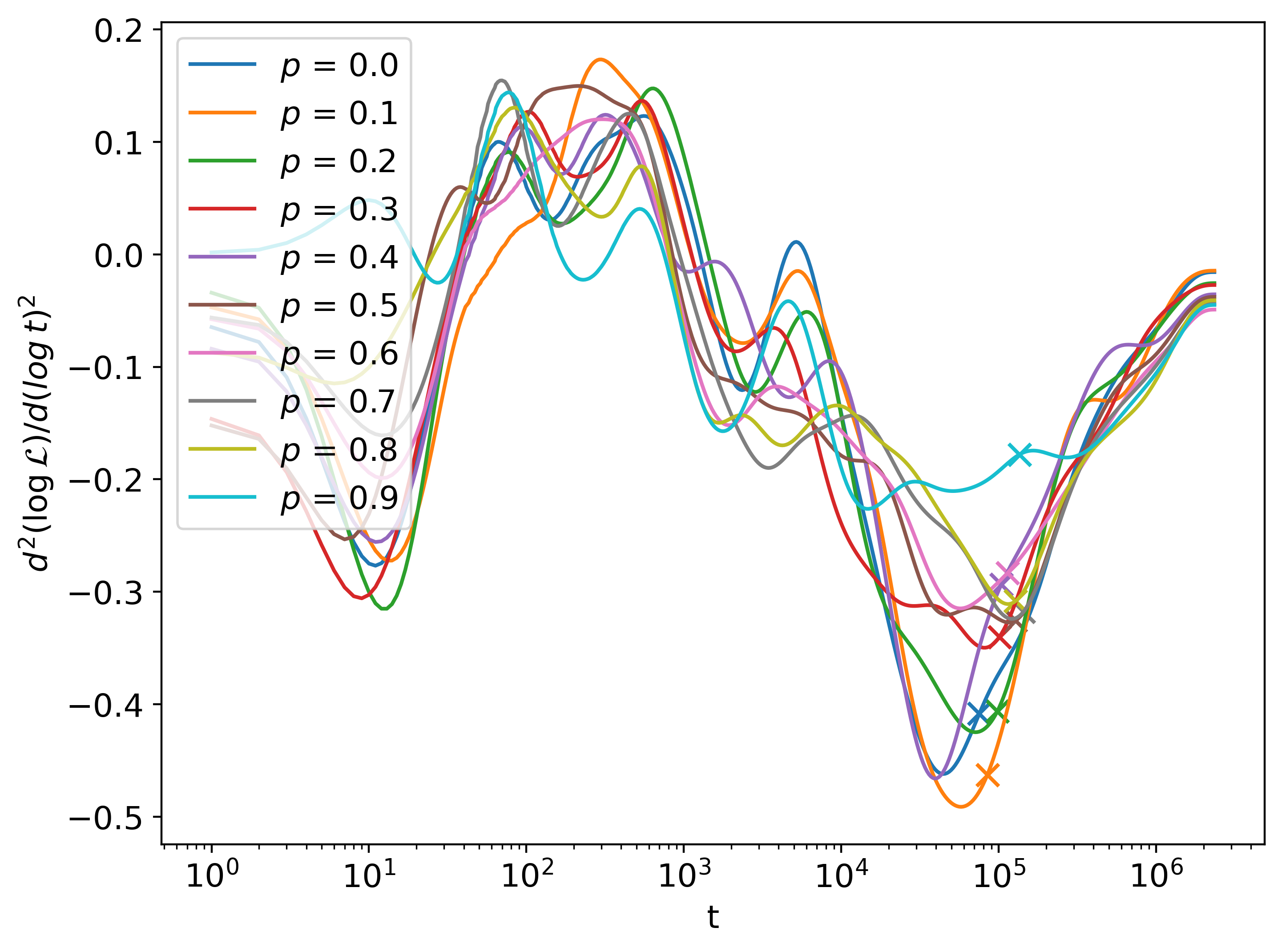}
  \end{subfigure}
  \hfill
  \begin{subfigure}[b]{0.49\textwidth}
    \includegraphics[width=\textwidth]{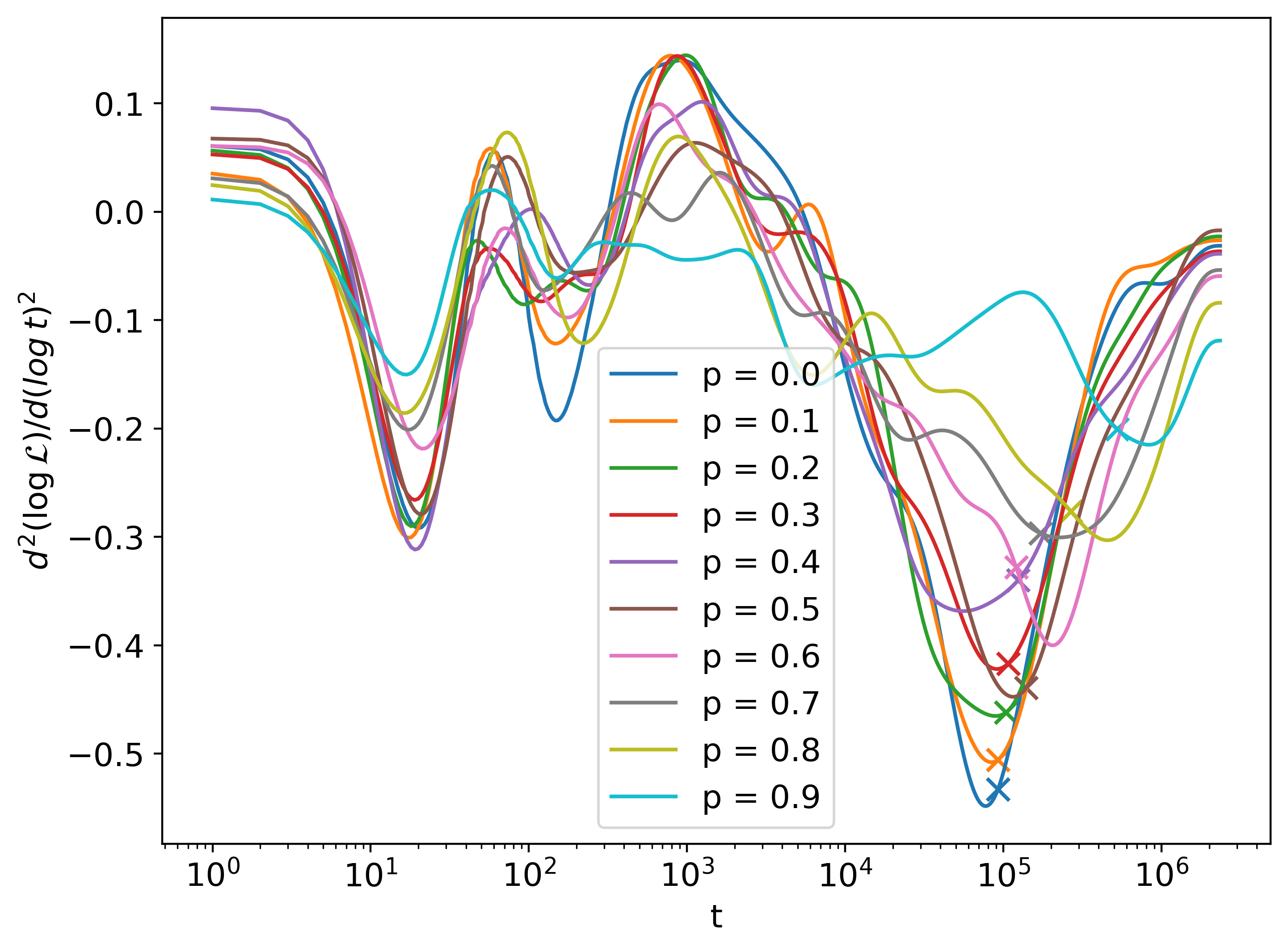}
  \end{subfigure}
  \caption{2\textsuperscript{nd}  derivative of $\log {\mathcal{L}}$ w.r.t $\log {t}$ for \textbf{(left)} SGD  and \textbf{(right)} Adadelta optimizers with learning rate 0.013 and different dropout probabilities on MNIST dataset. The symbol x represents the minimum of the test loss.}
  \label{fig:dropout_deriv}
\end{figure}

The trend in overfitting seen in Fig. \ref{fig:sgd_dropout_loss} is also reflected in Fig. \ref{fig:dropout_deriv}, where depth of the minima of $d^2(\log \mathcal{L}) / d(\log t)^2$, corresponding to overfitting, decreases with increasing dropout probability. Note that this trend is not perfect since it is subject to statistical fluctuations depending on the random initialization of parameters in the neural network. For example in Fig. \ref{fig:dropout_deriv} \textbf{(left)}, $p=0.1$ has a lower minima compared to $p=0.0$.

A plausible explanation for the trend in $d^2(\log \mathcal{L}) / d(\log t)^2$ is that, when dropout regularization is used, approximately $p$ fraction of the neurons are turned off at each training step and does not get updated. Therefore, fewer parameters are updated when $p$ is larger. This leads to smaller changes in the probability distribution of parameters and hence smaller changes to $\mathcal{L}$ with time. Based on this reasoning we expect to see lower $\mathcal{L}$ for higher values of $p$. However, Fig.~\ref{fig:mnist_dropout_il} shows that the actual trend is the opposite, with higher values of $p$ having larger $\mathcal{L}$. The simple explanation thus fails to account for the trends in $d^2(\log \mathcal{L}) / d(\log t)^2$ and warrants further study.

\FloatBarrier
\subsection{Noise Regularization}
\label{sec:noise}
\begin{figure}[h]
  \begin{subfigure}[b]{0.49\textwidth}
    \includegraphics[width=\textwidth]{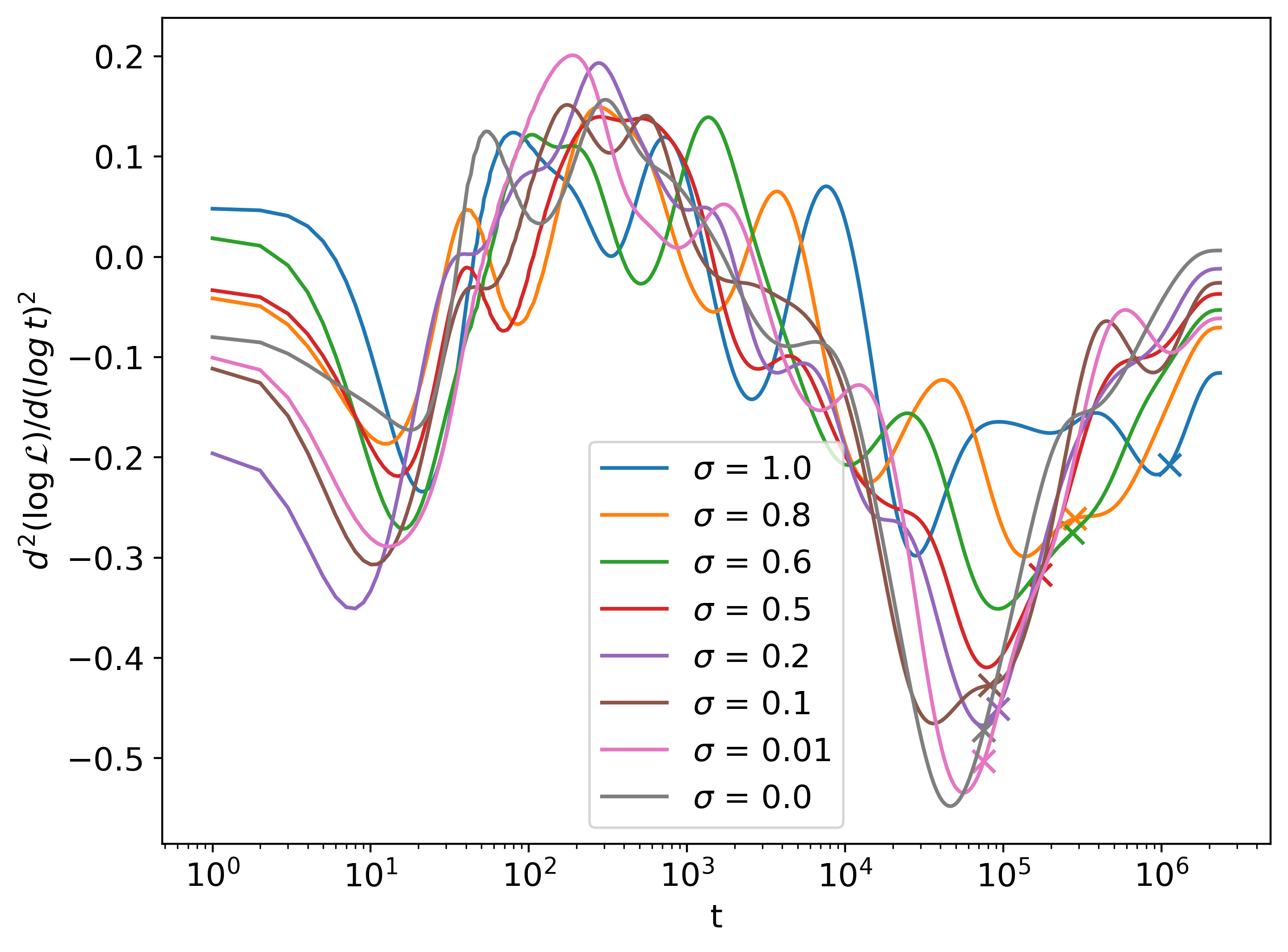}
  \end{subfigure}
  \hfill
  \begin{subfigure}[b]{0.49\textwidth}
    \includegraphics[width=\textwidth]{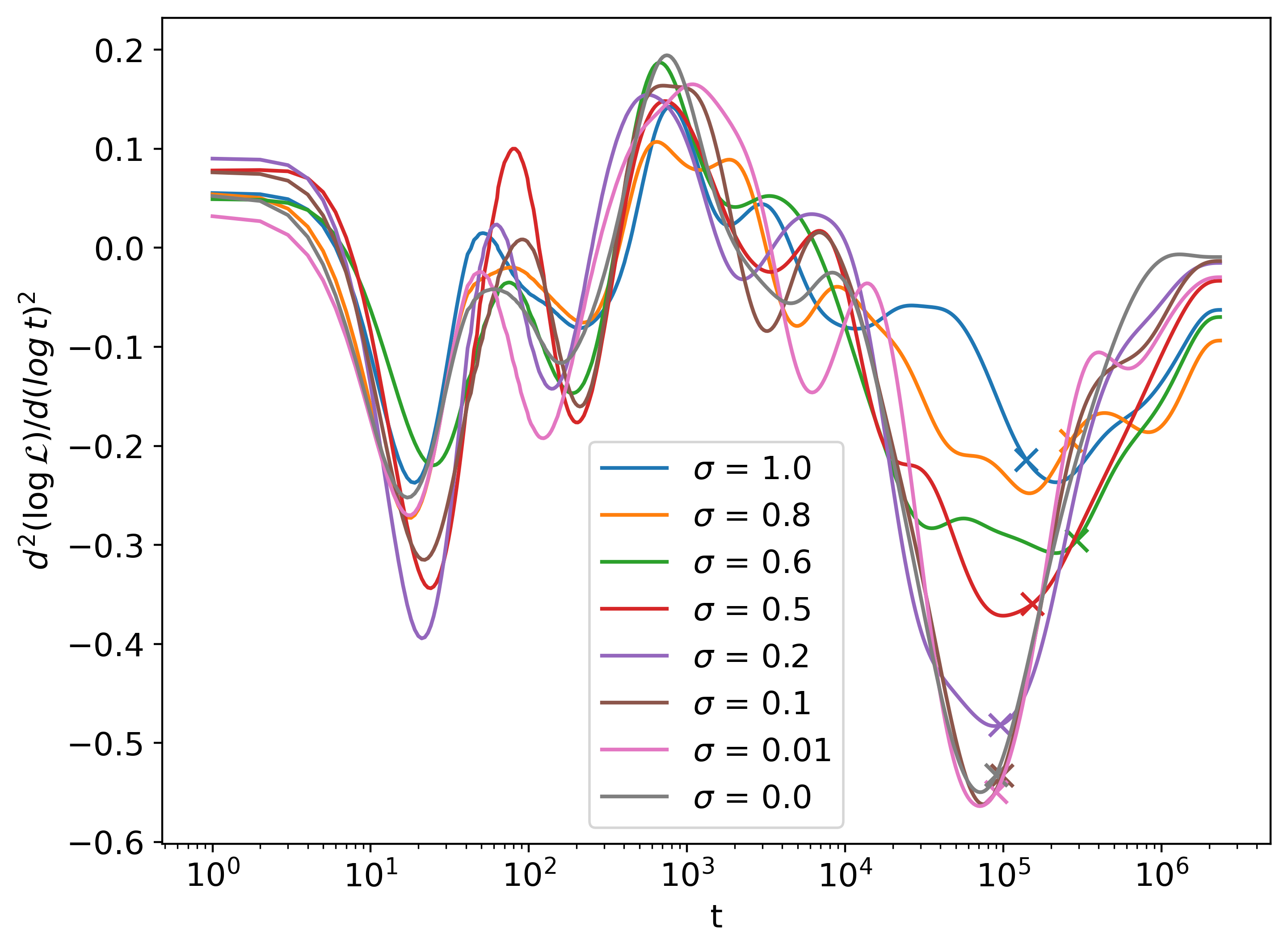}
  \end{subfigure}
  \caption{2\textsuperscript{nd}  derivative of $\log {\mathcal{L}}$ w.r.t $\log {t}$ for \textbf{(left)} SGD  and \textbf{(right)} Adadelta optimizers with learning rate 0.013 and augmentation with noise of different standard deviation values on MNIST dataset. The symbol 'x' represents the minimum of the test loss.}
  \label{fig:noise_deriv}
\end{figure}

Adding random noise to the data while training is another method of regularization commonly used in deep learning to learn more robust features and improve generalization. Noise regularization shows similar trends in the value of $d^2(\log \mathcal{L}) / d(\log t)^2$ as shown by dropout regularization. The higher the noise level, the lesser the extent of overfitting, the shallower the minima corresponding to overfitting. This can be seen in Fig.~\ref{fig:noise_deriv} for the case of SGD and Adadelta on MNIST dataset.
\FloatBarrier

\subsection{Fashion-MNIST}
\label{sec:fmnist}
\FloatBarrier

\begin{figure}[h]
	\begin{subfigure}[b]{0.49\textwidth}
		\includegraphics[width=\textwidth]{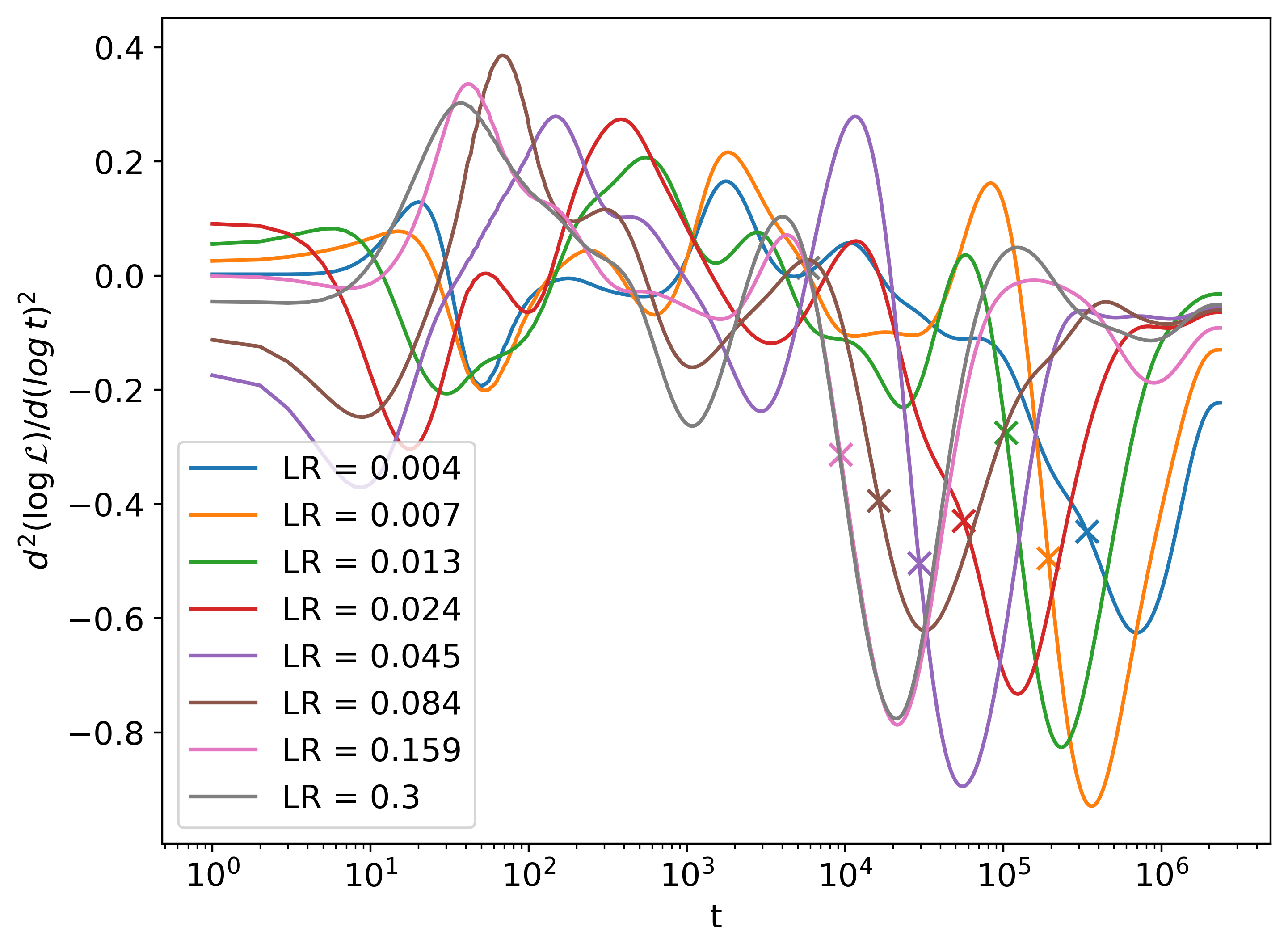}
	\end{subfigure}
	\hfill
	\begin{subfigure}[b]{0.49\textwidth}
		\includegraphics[width=\textwidth]{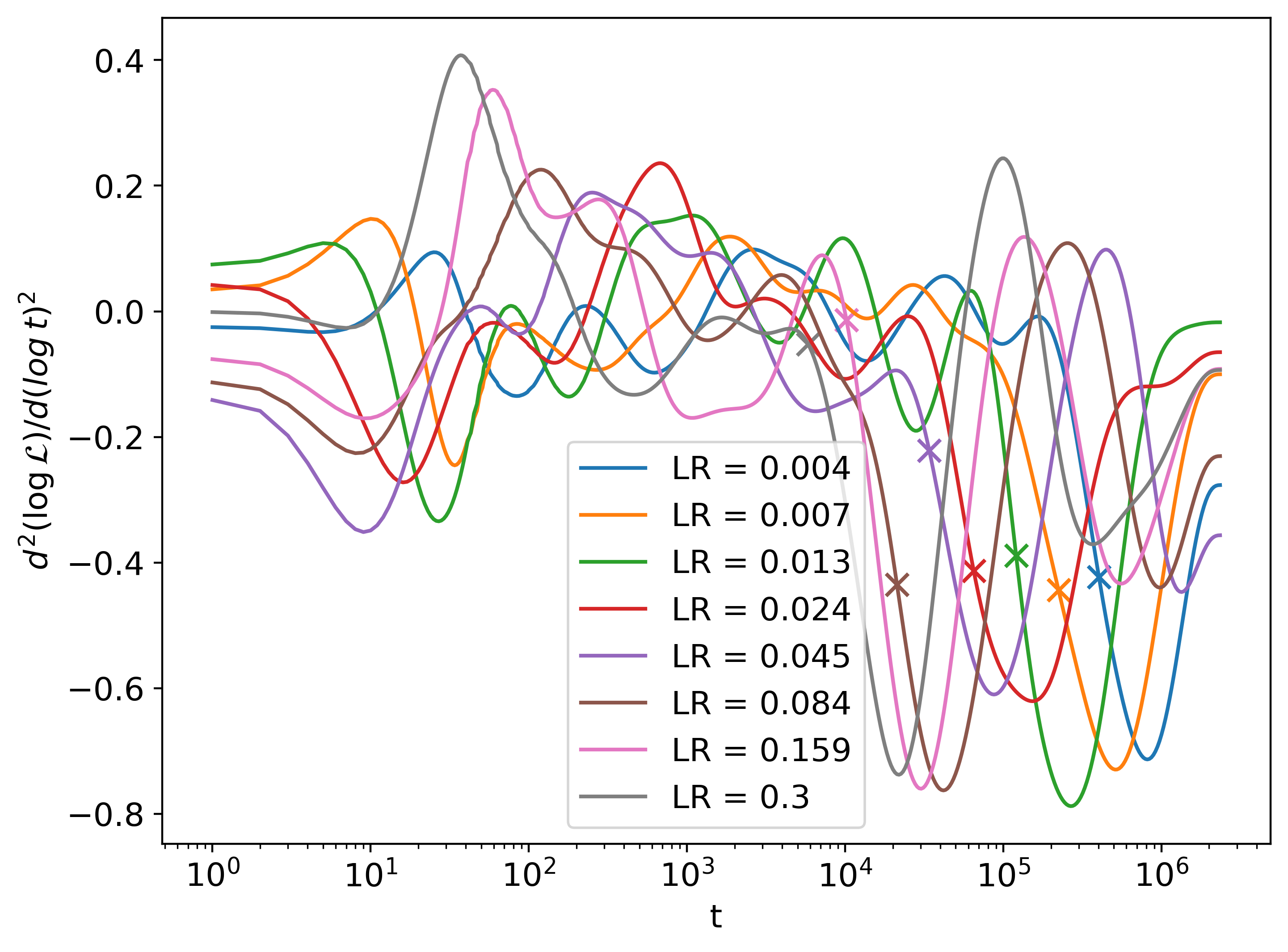}
	\end{subfigure}
	\caption{2\textsuperscript{nd}  derivative of $\log {\mathcal{L}}$ w.r.t $\log {t}$ for \textbf{(left)} SGD  and \textbf{(right)} Adadelta with different learning rates on Fashion-MNIST dataset. The symbol 'x' represents the minimum of the test loss.}
	\label{fig:fmnist_lr_deriv}
\end{figure}

\begin{figure}[h]
	\begin{subfigure}[b]{0.49\textwidth}
		\includegraphics[width=\textwidth]{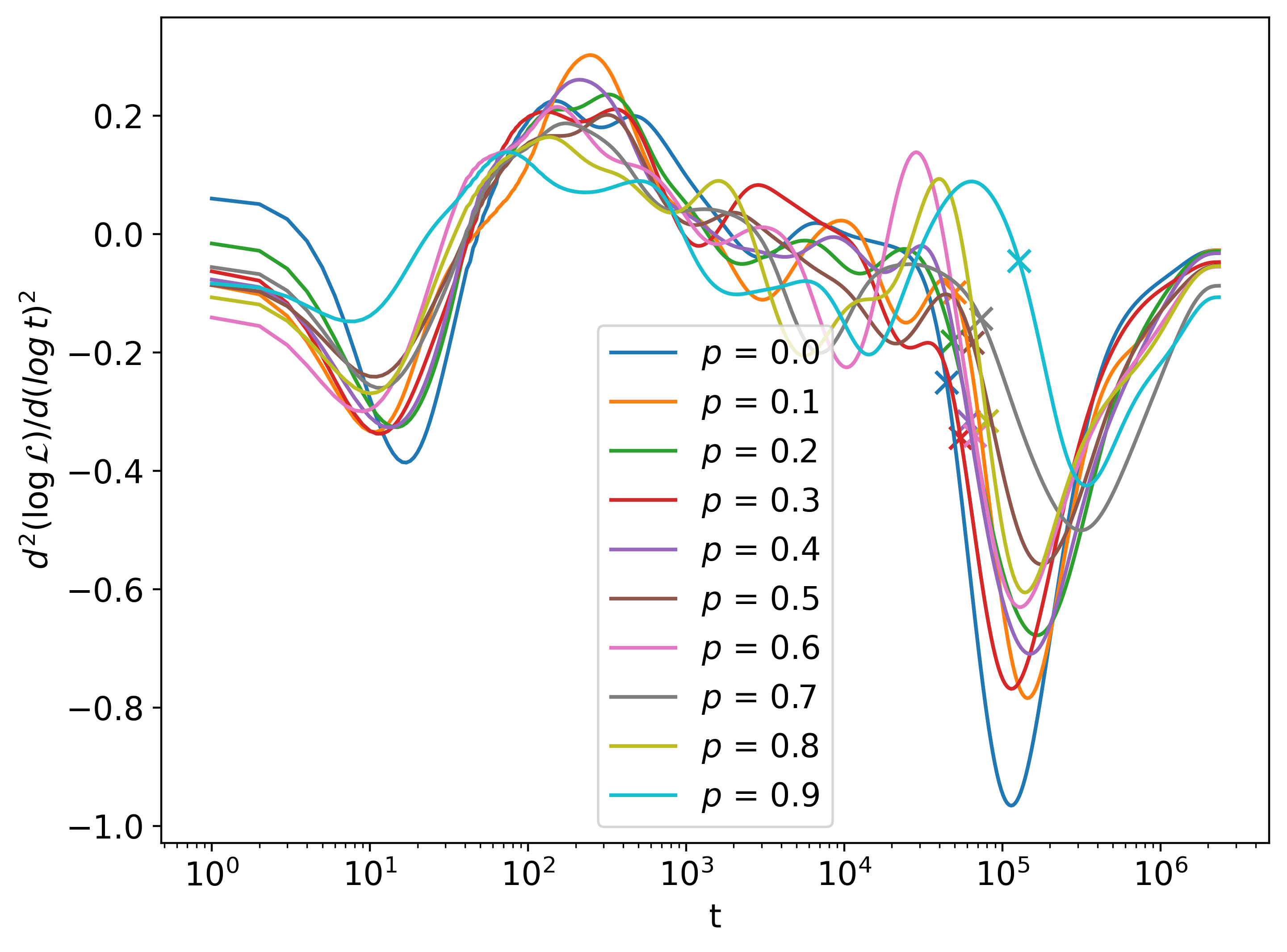}    
	\end{subfigure}
	\hfill
	\begin{subfigure}[b]{0.49\textwidth}
		\includegraphics[width=\textwidth]{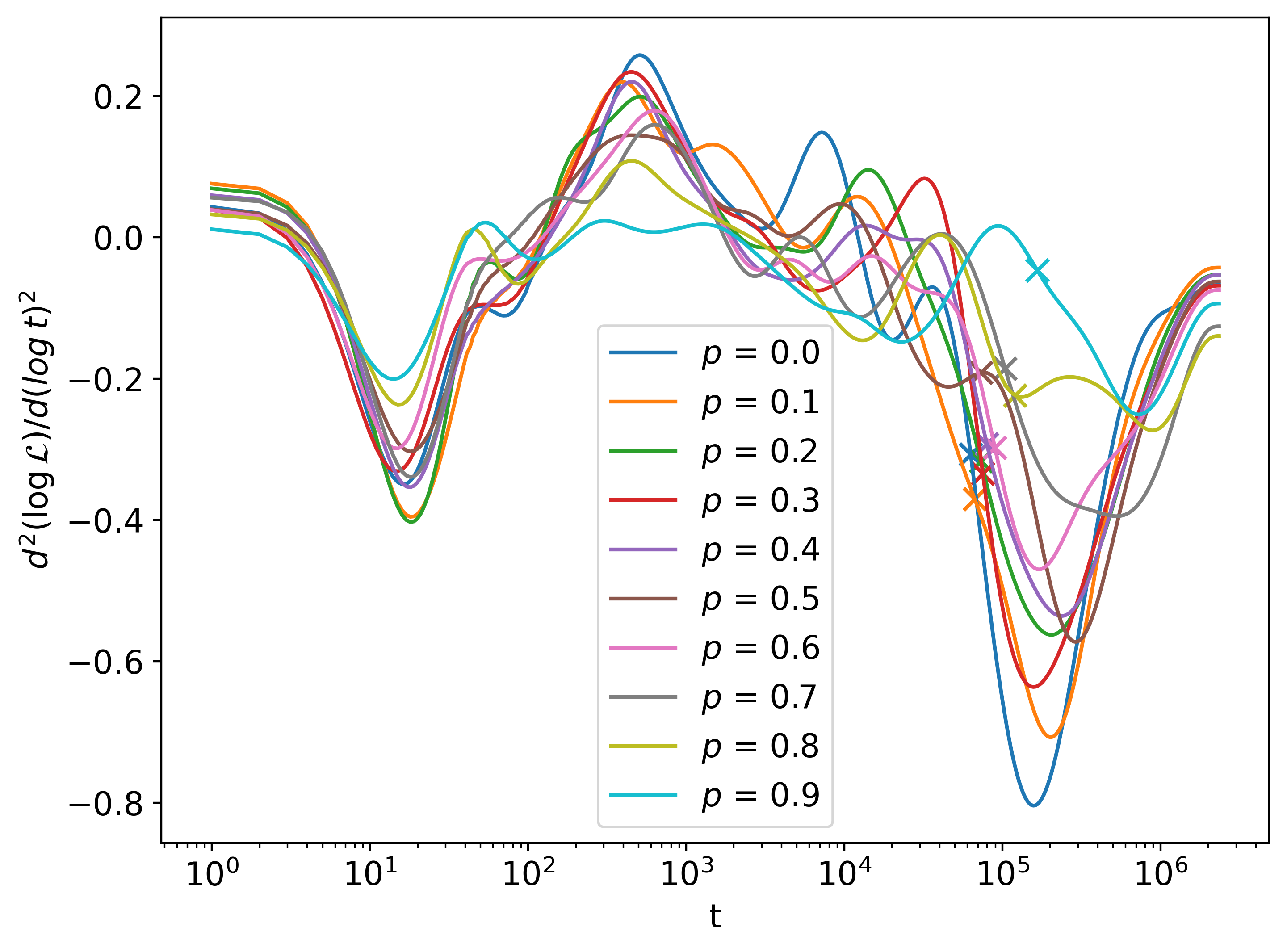}
	\end{subfigure}
	\caption{2\textsuperscript{nd}  derivative of $\log {\mathcal{L}}$ w.r.t $\log {t}$ for \textbf{(left)} SGD  and \textbf{(right)} Adadelta optimizers with learning rate 0.013 and different dropout probabilities on Fashion-MNIST dataset. The symbol 'x' represents the minimum of the test loss.}
	\label{fig:fmnist_dropout_deriv}
\end{figure}

\begin{figure}[h]
  \begin{subfigure}[b]{0.49\textwidth}
    \includegraphics[width=\textwidth]{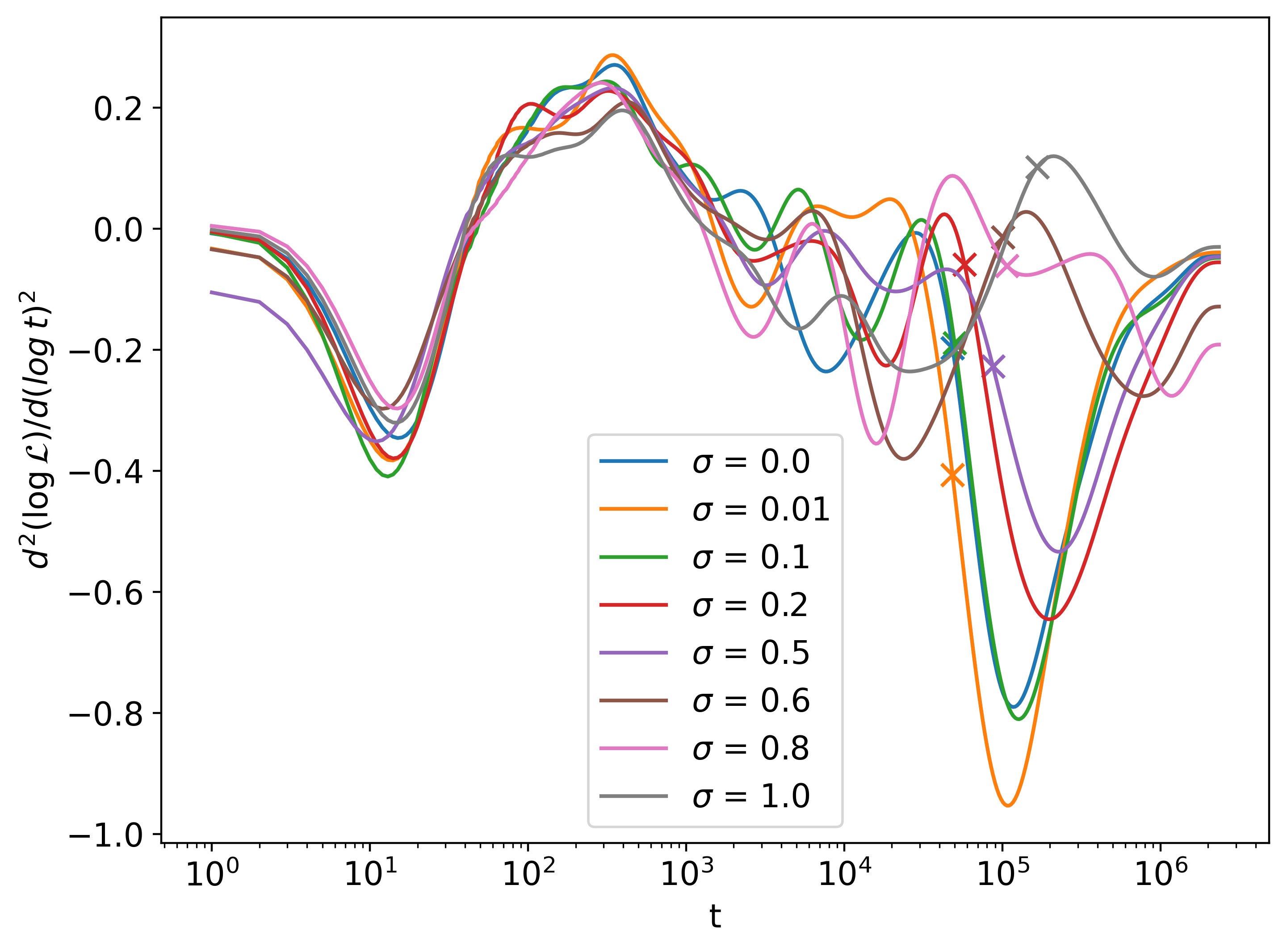}
  \end{subfigure}
  \hfill
  \begin{subfigure}[b]{0.49\textwidth}
    \includegraphics[width=\textwidth]{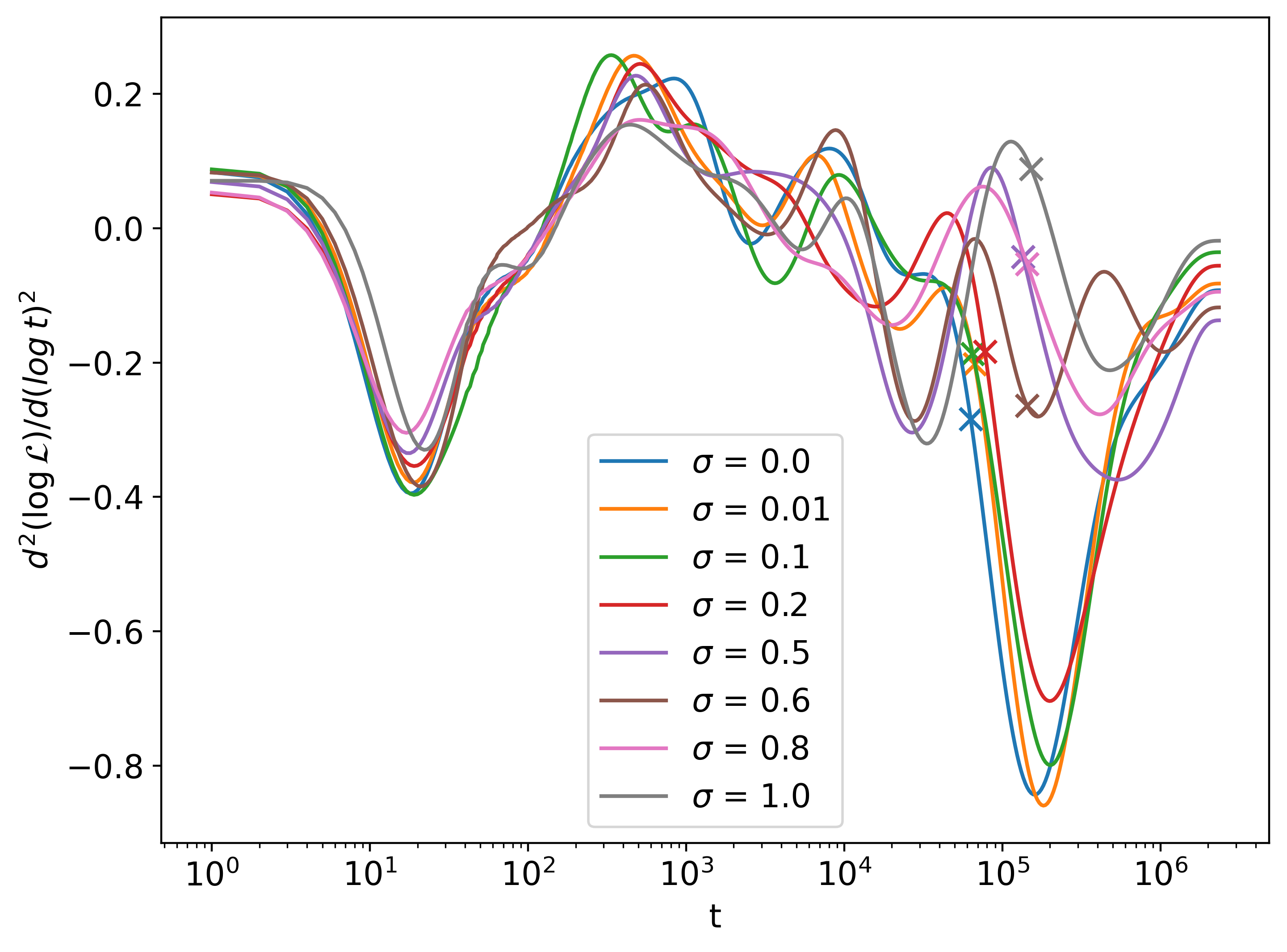}
  \end{subfigure}
  \caption{2\textsuperscript{nd}  derivative of $\log {\mathcal{L}}$ w.r.t $\log {t}$ for \textbf{(left)} SGD  and \textbf{(right)} Adadelta optimizers with learning rate 0.013 and augmentation with noise of different standard deviation values on Fashion-MNIST dataset. The symbol 'x' represents the minimum of the test loss.}
  \label{fig:fmnist_noise_deriv}
\end{figure}

The trends discussed in Sec. \ref{sec:lr}, Sec. \ref{sec:dropout} and Sec. \ref{sec:noise} can also be seen with the Fashion-MNIST dataset. The difference here is that, for the MNIST dataset, the minima of $d^2(\log \mathcal{L}) / d(\log t)^2$ precede the minima of the test loss, however, for Fashion-MNIST, the minima of $d^2(\log \mathcal{L}) / d(\log t)^2$ come after the minima of the test loss, as can be seen in Fig. \ref{fig:fmnist_lr_deriv}, Fig. \ref{fig:fmnist_dropout_deriv} and Fig. \ref{fig:fmnist_noise_deriv}. This can be understood by considering that overfitting is represented better by a finite interval in the training curve rather than a minimum point which can shift for different training runs. In the case of Fashion-MNIST this region is much wider and noisier (Fig. \ref{fig:fmnist_test_loss}) compared to MNIST (Fig. \ref{fig:mnist_lr_test_loss}).

The trends in dropout regularization and noise regularization discussed in Sec. \ref{sec:dropout} and Sec. \ref{sec:noise}, respectively, hold true for the case of Fashion-MNIST. This is shown in Fig. \ref{fig:fmnist_dropout_deriv} for the case of dropout regularization and Fig. \ref{fig:fmnist_noise_deriv} for noise regularization.

\FloatBarrier

{
\subsection{ResNet CIFAR-10}
\label{sec:resnet}
So far we have looked at the behavior of a shallow fully connected neural network under different hyperparameters. Further evidence of the generality of this behavior can be seen by looking at a more complex neural network. For this we use the ResNet-50 \cite{he2016deep} model which is a convolutional neural network with 50 layers. The model was trained on the CIFAR-10 dataset. The ResNet-50 is a fairly large neural network with around 25 million parameters. Since the number of parameters is much larger than the size of the dataset we see significant overfitting, which can be observed in Fig. \ref{fig:resnet} \textbf{(left)}. 

\begin{figure}[h]
  \begin{subfigure}[b]{0.49\textwidth}
    \includegraphics[width=\textwidth]{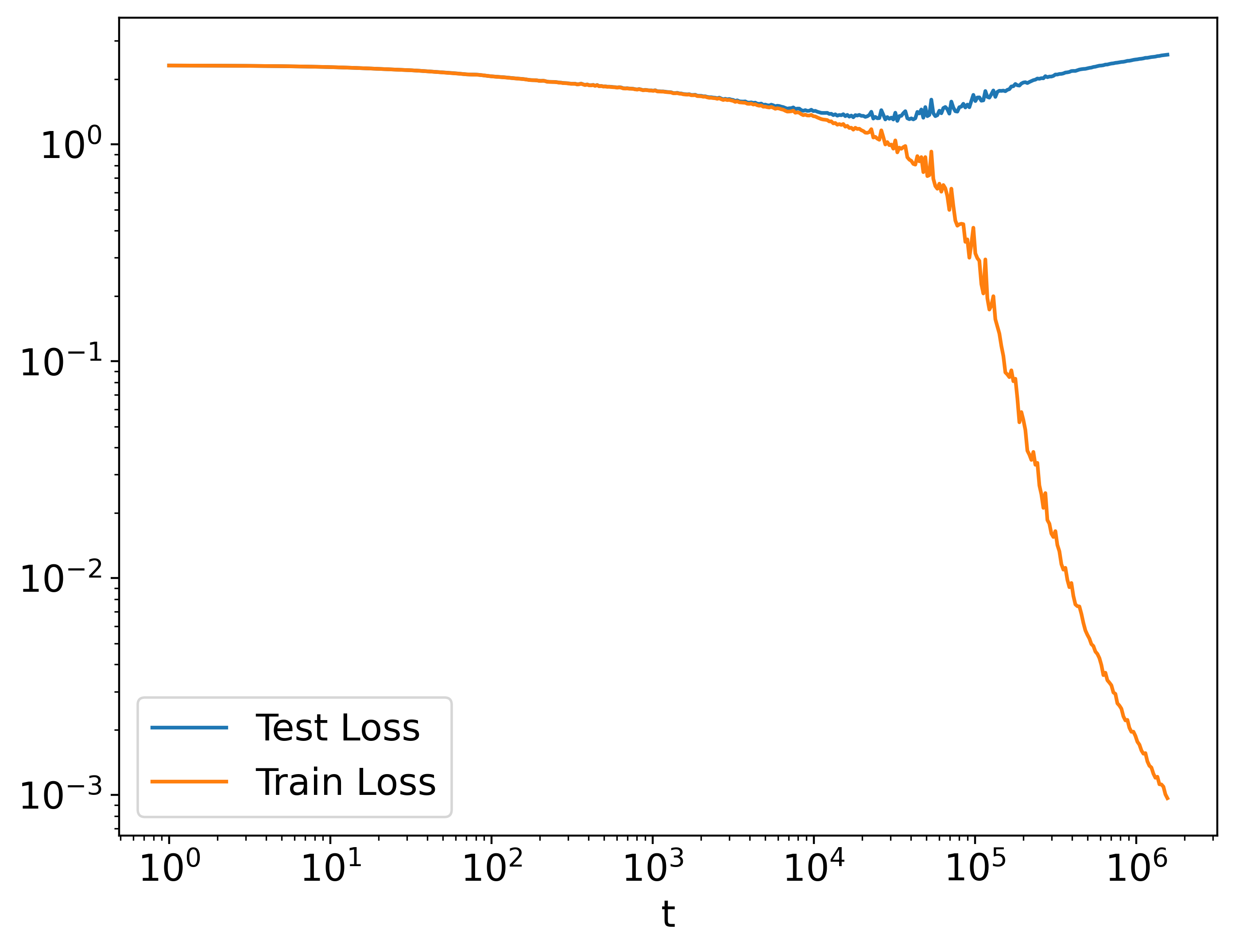}
  \end{subfigure}
  \hfill
  \begin{subfigure}[b]{0.49\textwidth}
    \includegraphics[width=\textwidth]{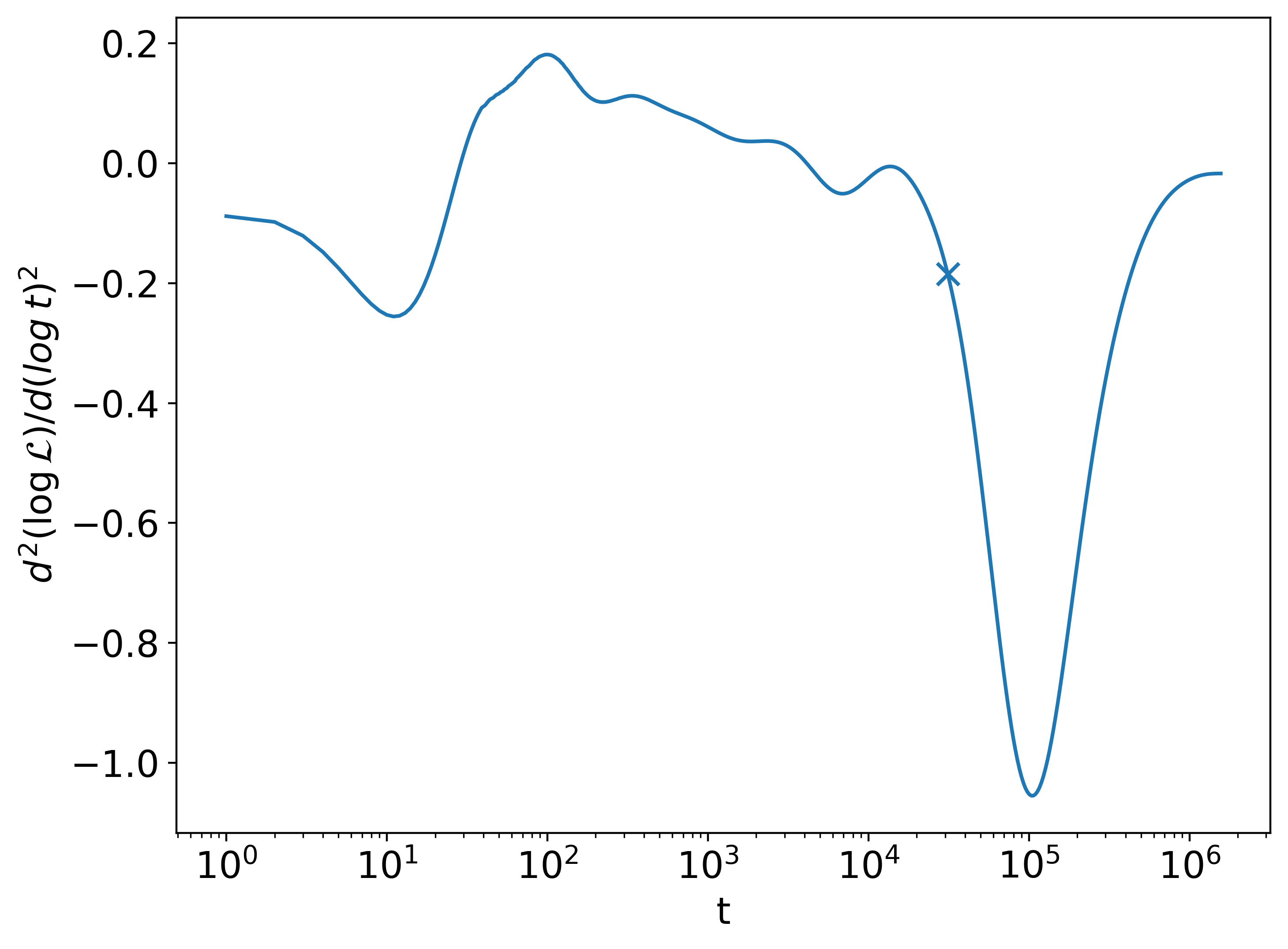}
  \end{subfigure}
  \caption{\textbf{(left)} Train and test loss for ResNet-50 trained on CIFAR-10 dataset using SGD optimizer with learning rate 0.013. The corresponding \textbf{(right)} 2\textsuperscript{nd}  derivative of $\log {\mathcal{L}}$ w.r.t $\log {t}$. The symbol 'x' represents the minimum of the test loss.}
  \label{fig:resnet}
\end{figure}

In Fig. \ref{fig:resnet} \textbf{(right)}, similar to the case of Fashion-MNIST the minima of the test loss appears before the minima of $d^2(\log \mathcal{L}) / d(\log t)^2$. This is explained by the much more noisy overfitting range compared to the case of shallow network trained on MNIST. A zoomed-in view of the test loss for ResNet-50 is given in Fig. \ref{fig:resnet_test_loss}. Note that compared to the shallow network the 2\textsuperscript{nd}  derivative of ResNet-50 exhibits a deeper valley. This can be understood in terms of stronger overfitting as discussed in the case of dropout and noise regularization, as well as from the point of view of finite-size scaling discussed in Sec. \ref{sec:il_phase_transition}, since ResNet-50 has a much larger number of parameters compared to the shallow network previously considered.
}

\section{Analogy to Phase Transition}
\label{sec:il_phase_transition}
In statistical physics the Boltzmann distribution for the probability $p_i$ of a system to be in state $i$ with energy $E_i$ is given by
\begin{equation}
    p_i=\frac{1}{Z} e^{-\beta E_i},\ \beta=1 / k_B T.
\end{equation}
Here the normalizing factor $Z$ is called the partition function, $k_B$ is the Boltzmann constant and $T$ is the temperature. The normalization factor can be defined as $Z = \sum_i e^{-\beta E_i}$. Then the expectation value of energy is given by
\begin{equation}
    \langle E(\beta) \rangle=-\frac{d \ln Z}{d \beta},
    \label{eq:exp_energy}
\end{equation}
and the variance of energy or energy fluctuation is given by 
\begin{equation}
    \left\langle(E-\langle E\rangle)^2\right\rangle=\frac{\partial^2 \ln Z}{\partial \beta^2}.
    \label{eq:var_energy}
\end{equation}

In thermodynamics a first order phase transition is marked by a discontinuity in the expected energy function or the first derivative of $\log z$ given by Eq. \ref{eq:exp_energy},  whereas a continuous phase transition has a continuous expected energy but a diverging energy fluctuation or the second derivative of $\log z$ given by Eq. \ref{eq:var_energy}. Note that the these statements are valid in the thermodynamic limit where the system has infinite size. For systems of finite size, a steep slope instead of a discontinuity in the expected value of energy will be observed, and the divergence in the energy fluctuation will become a large peak or a valley. This behavior is seen throughout Sec. \ref{sec:exp_res} in the quantity $d^2(\log \mathcal{L}) / d(\log t)^2$.

\begin{figure}[h]
  \begin{subfigure}[b]{0.49\textwidth}
    \includegraphics[width=\textwidth]{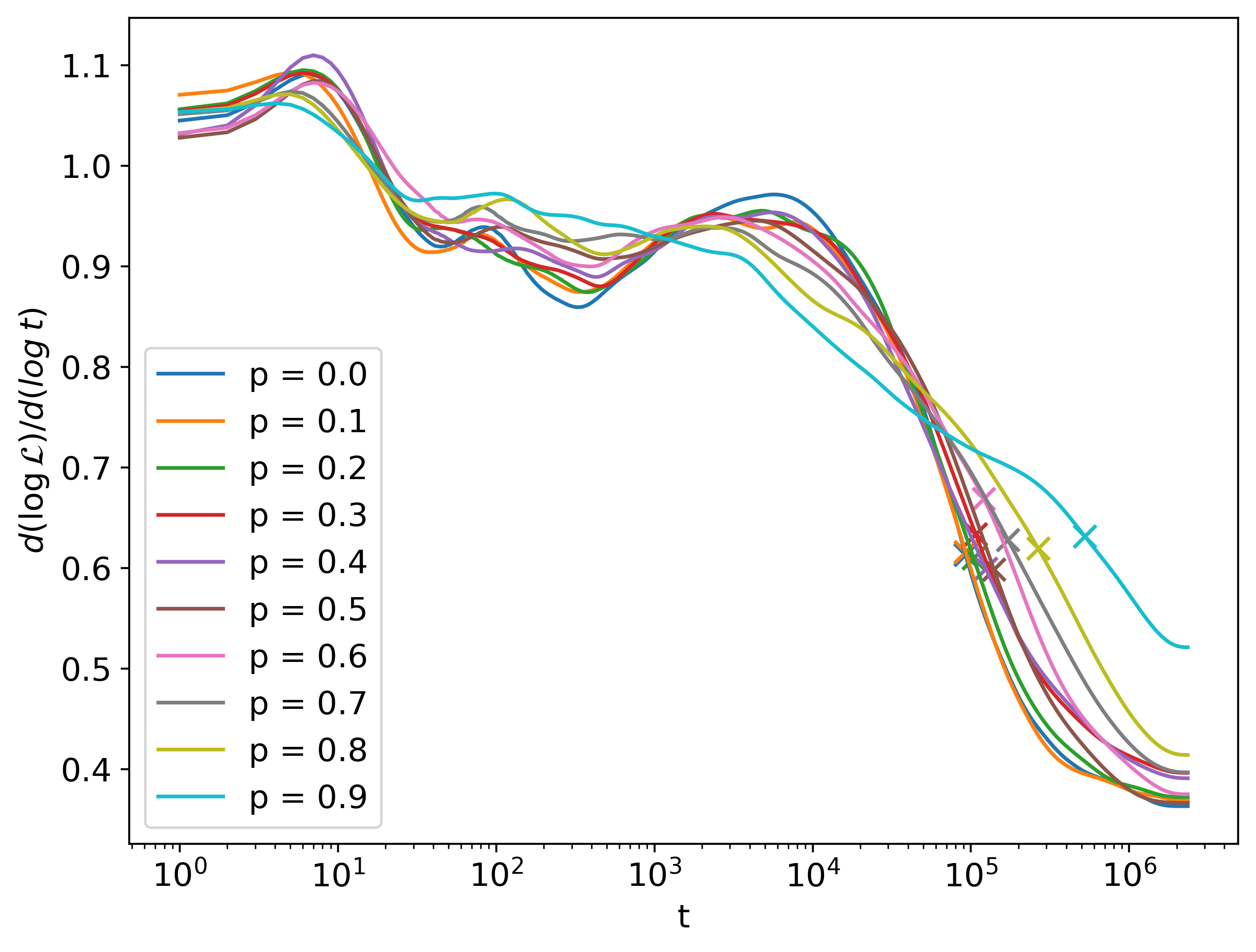}
  \end{subfigure}
  \hfill
  \begin{subfigure}[b]{0.49\textwidth}
    \includegraphics[width=\textwidth]{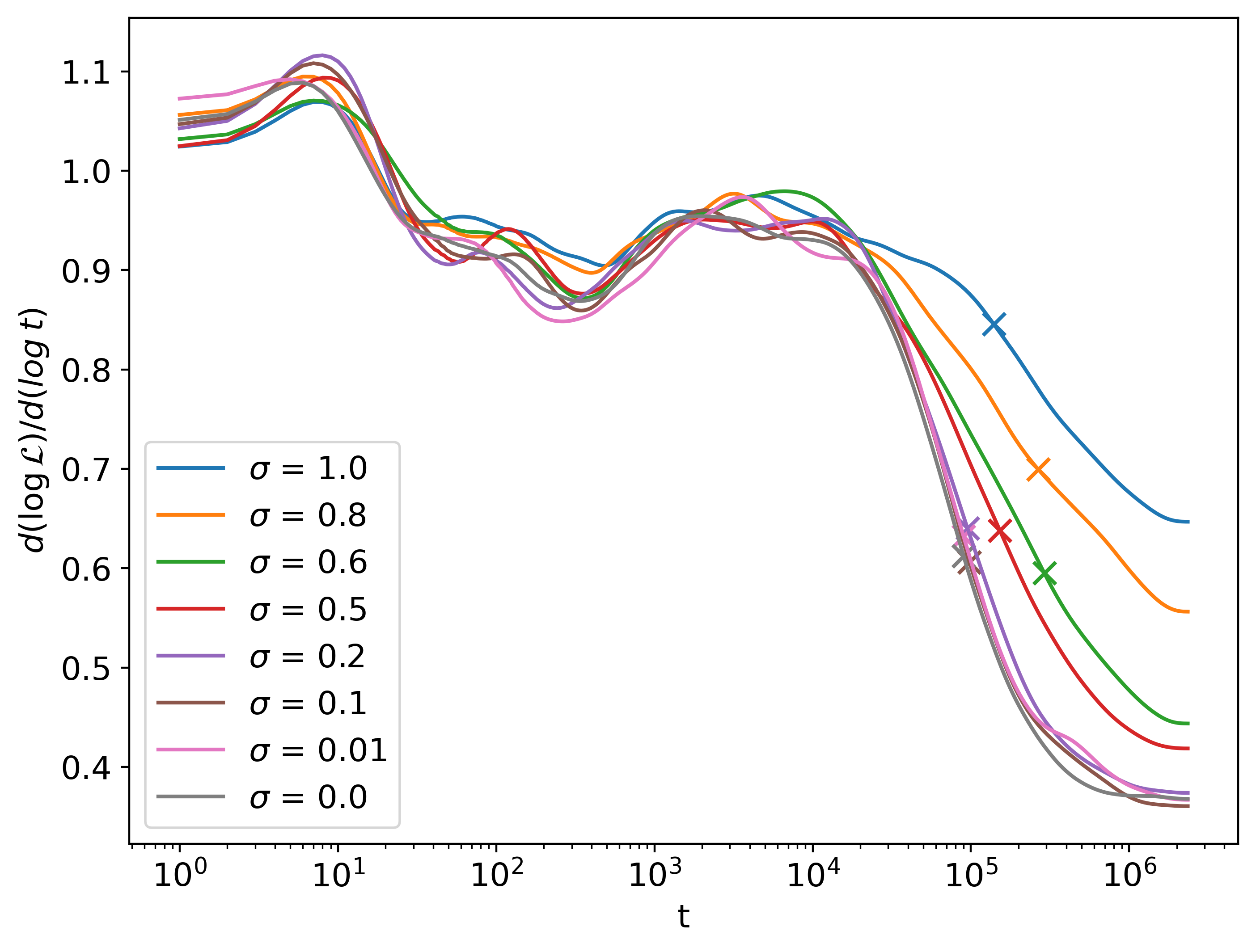}
  \end{subfigure}
  \caption{Derivatives of $\log {\mathcal{L}}$ w.r.t $\log {t}$ for Adaldelta with lr = 0.013 with different \textbf{(left)} dropout and \textbf{(right)} noise levels trained on the MNIST dataset. The symbol x represents the minimum of the test loss.}
  \label{fig:adadelta_dropout_std_deriv1}
\end{figure}

The finite-size scaling-like behavior is more apparent in the case of dropout and noise regularization. For the first derivatives (Fig. \ref{fig:adadelta_dropout_std_deriv1}) the smaller the values of dropout and noise, the steeper the slopes near over-fitting. For the second derivatives (Fig. \ref{fig:dropout_deriv}, \ref{fig:noise_deriv}, \ref{fig:fmnist_dropout_deriv}, \ref{fig:fmnist_noise_deriv}), deeper valleys are seen with smaller values of dropout and noise. The reason for this is the system size decreases if the dropout probability increases. Furthermore, with increased noise, the training data is coarse-grained, in a sense decreasing the system size. More work is needed to make these statements concrete. Preliminary results given in Fig.~\ref{fig:adadelta_dropout_std_scaling} show that the slopes of Fig.~\ref{fig:adadelta_dropout_std_deriv1} at $t$ where test loss is minimum are linearly related to the standard deviation of the noise and dropout probability.

\section{Conclusion and Future Work}
\label{sec:conclusion_nn}
In this work, we demonstrated applicability of information geometry in the training of ANNs and studied how it captures the changes in the collective behavior of parameters during the training process. This information was then shown to forecast overfitting while training. In Sec. \ref{sec:exp_res} we saw different optimizers exhibiting a wide variety of behaviors on MNIST and Fashion-MNIST dataset. The focus of this study was on SGD and Adadelta optimizers which had the simplest behavior. While training with SGD and Adadelta optimizers, the information length initially scaled linearly with the number of training steps and changed its behavior when the model started overfitting. Using regularized derivatives described in Sec. \ref{sec:reg_deriv}, we were able to take the derivatives of the noisy information length data. The transition in the behavior of information length when the model was overfitting can be identified as a local minima in the quantity $d^2(\log \mathcal{L}) / d(\log t)^2$. Further investigation of regularization methods—dropout and noise augmentation—revealed that, as the extent of overfitting decreased, the depth of the local minima also decreased. These demonstrate that information length provides a useful tool to identify and forecast overfitting without referring to the test dataset.

In Sec. \ref{sec:il_phase_transition} we discussed how the behavior of 
$\frac{d(\log \mathcal{L})}{d(\log t)}$ and $\frac{d^2(\log \mathcal{L})}{d(\log t)^2}$ 
under over-fitting is analogous to the behavior of $\frac{\partial \ln Z}{\partial \beta}$ and $\frac{\partial^2 \ln Z}{\partial \beta^2}$ respectively, in a thermodynamic system undergoing a phase transition. Dropout and noise regularization is shown to exhibit finite-size scaling like behavior due to the change in the effective size of the system qualitatively. A more quantitative investigation into the critical exponents will be undertaken in the future.

Furthermore, it will be of interest to extend the analysis in this paper to investigate more complex behaviors exhibited by other optimizers and why Adadelta being an adaptive learning rate algorithm behaves similar to SGD as well as theoretical investigation into the universality of these behaviors.

\appendix
\section{Additional Figures}
\FloatBarrier
\begin{figure}[h]
  \begin{subfigure}[b]{0.49\textwidth}
    \includegraphics[width=\textwidth]{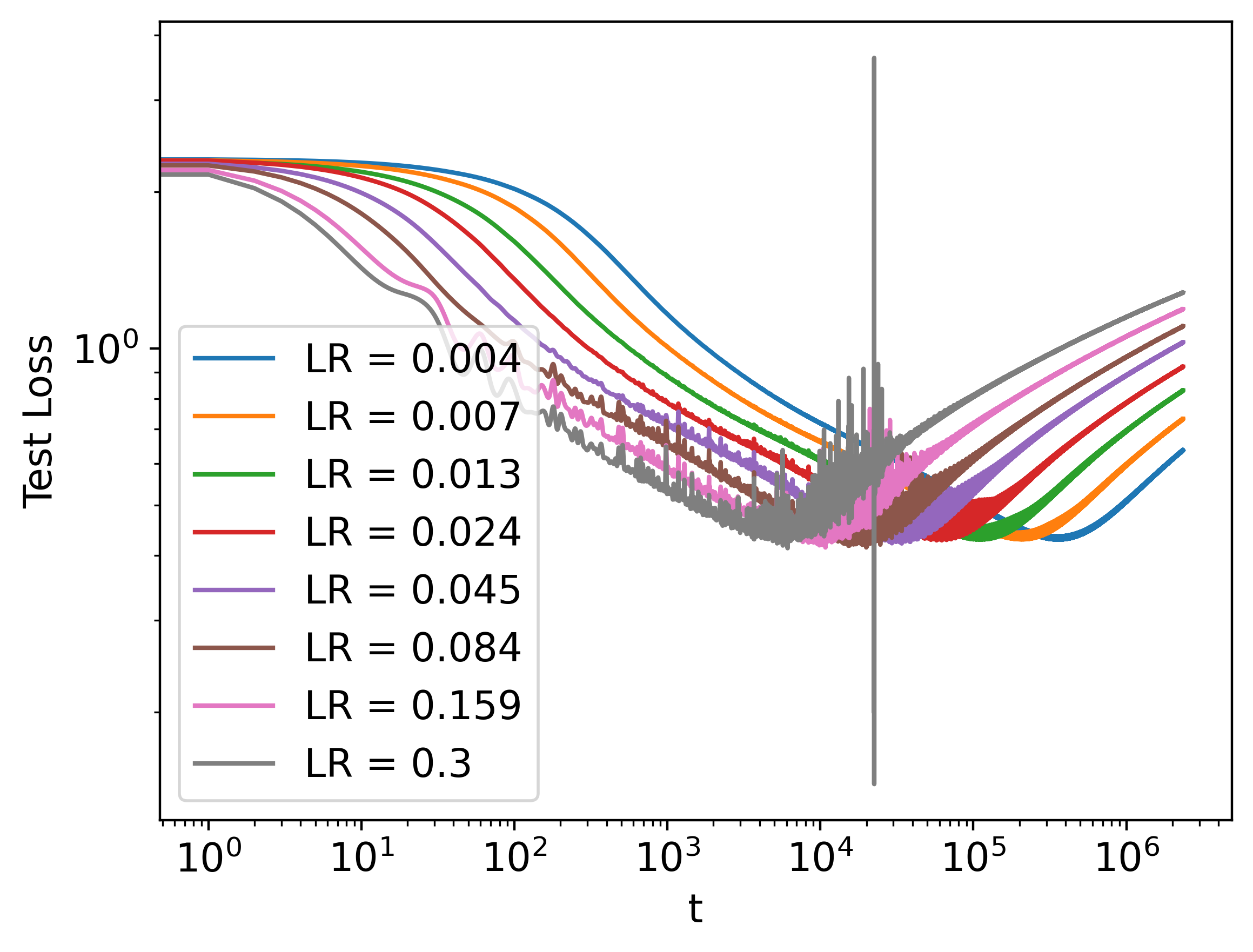}
  \end{subfigure}
  \hfill
  \begin{subfigure}[b]{0.49\textwidth}
    \includegraphics[width=\textwidth]{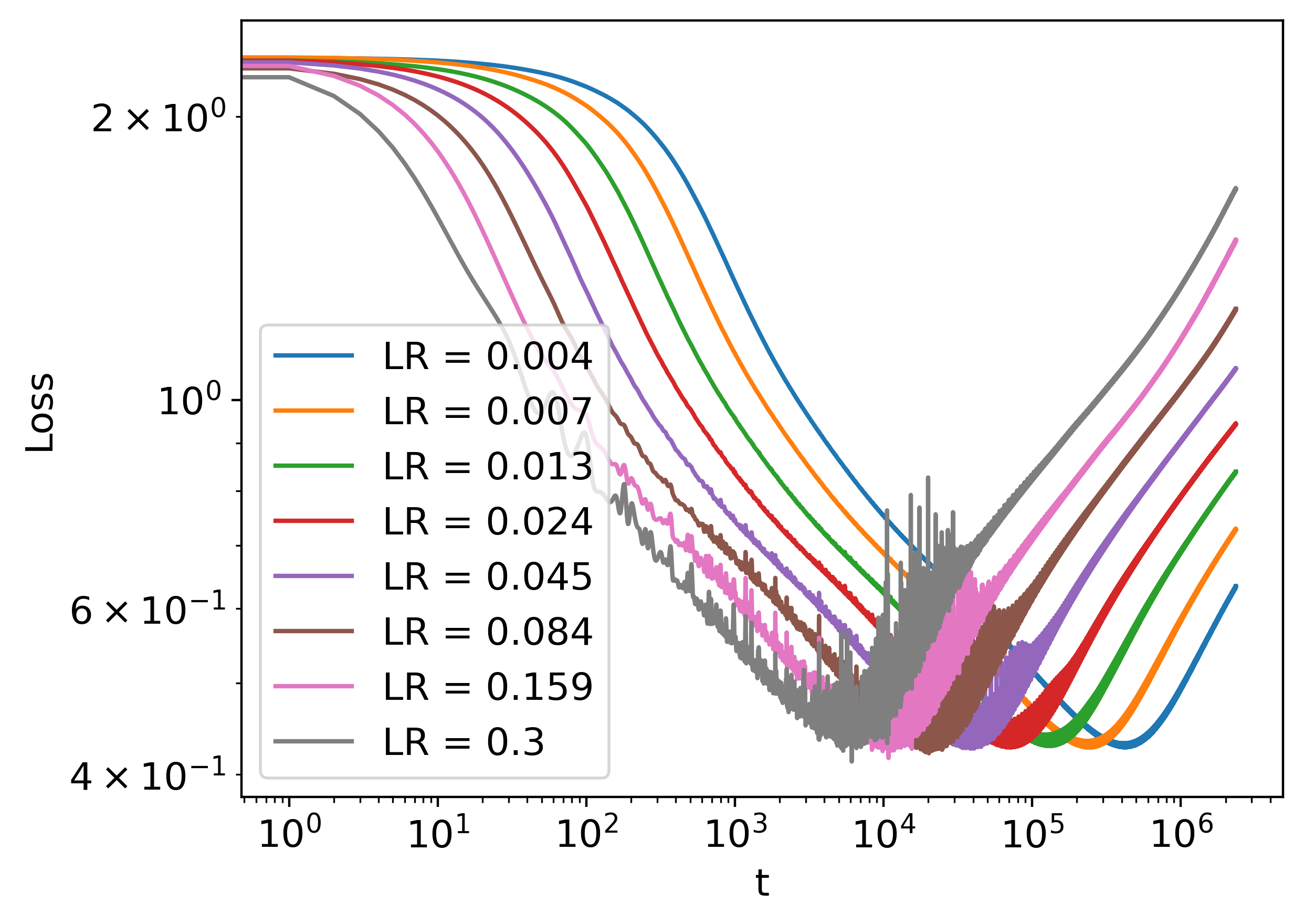}
  \end{subfigure}
  \caption{Loss on test data for \textbf{(left)} SGD and \textbf{(right)} Adadelta optimizers with different learning rates on Fashion-MNIST dataset. Note that the large spike in the case of learning rate 0.3 is a numerical artifact.}
  \label{fig:fmnist_test_loss}
\end{figure}

\begin{figure}[h]
  \begin{subfigure}[b]{0.49\textwidth}
    \includegraphics[width=\textwidth]{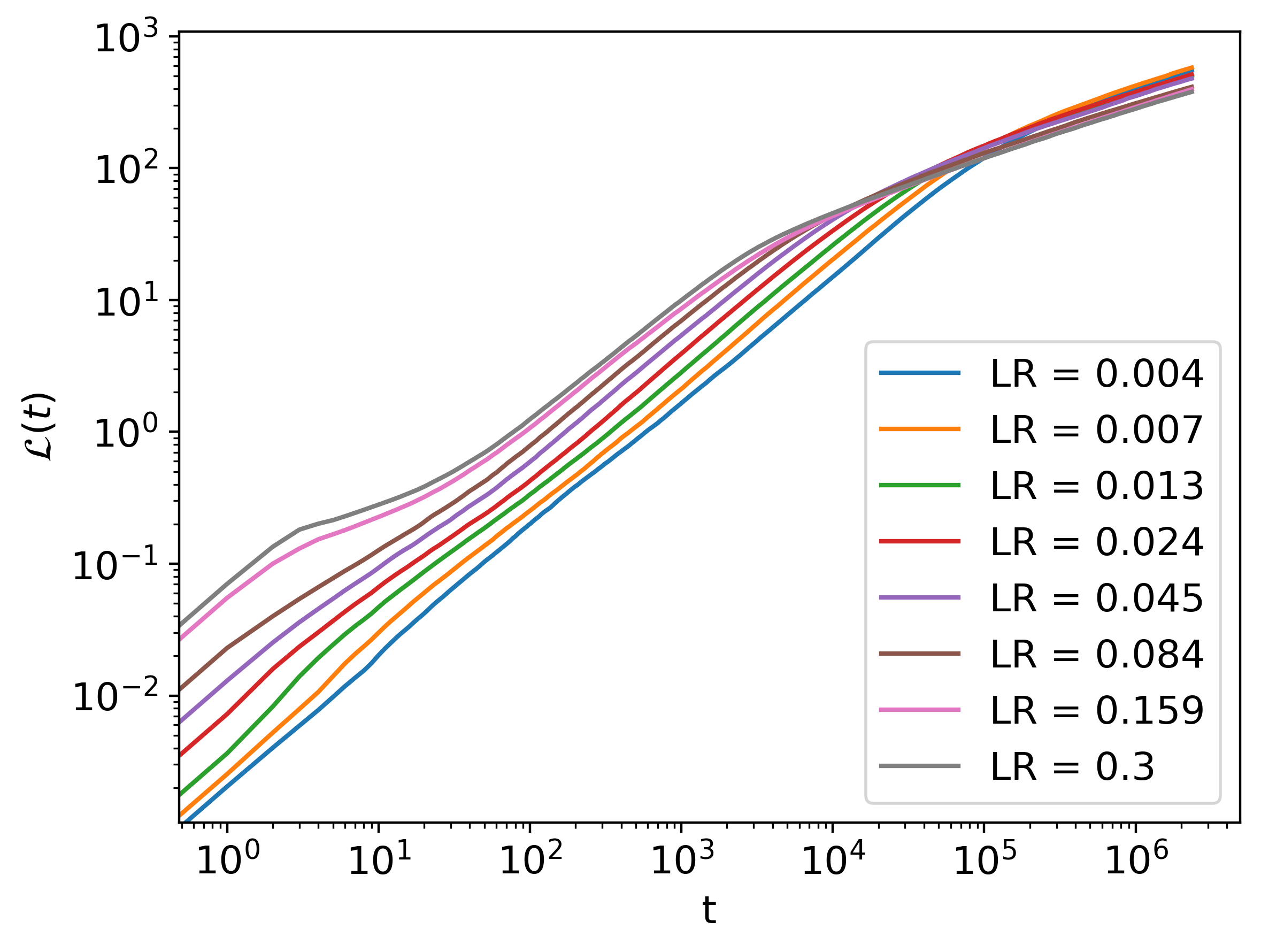}
  \end{subfigure}
  \hfill
  \begin{subfigure}[b]{0.49\textwidth}
    \includegraphics[width=\textwidth]{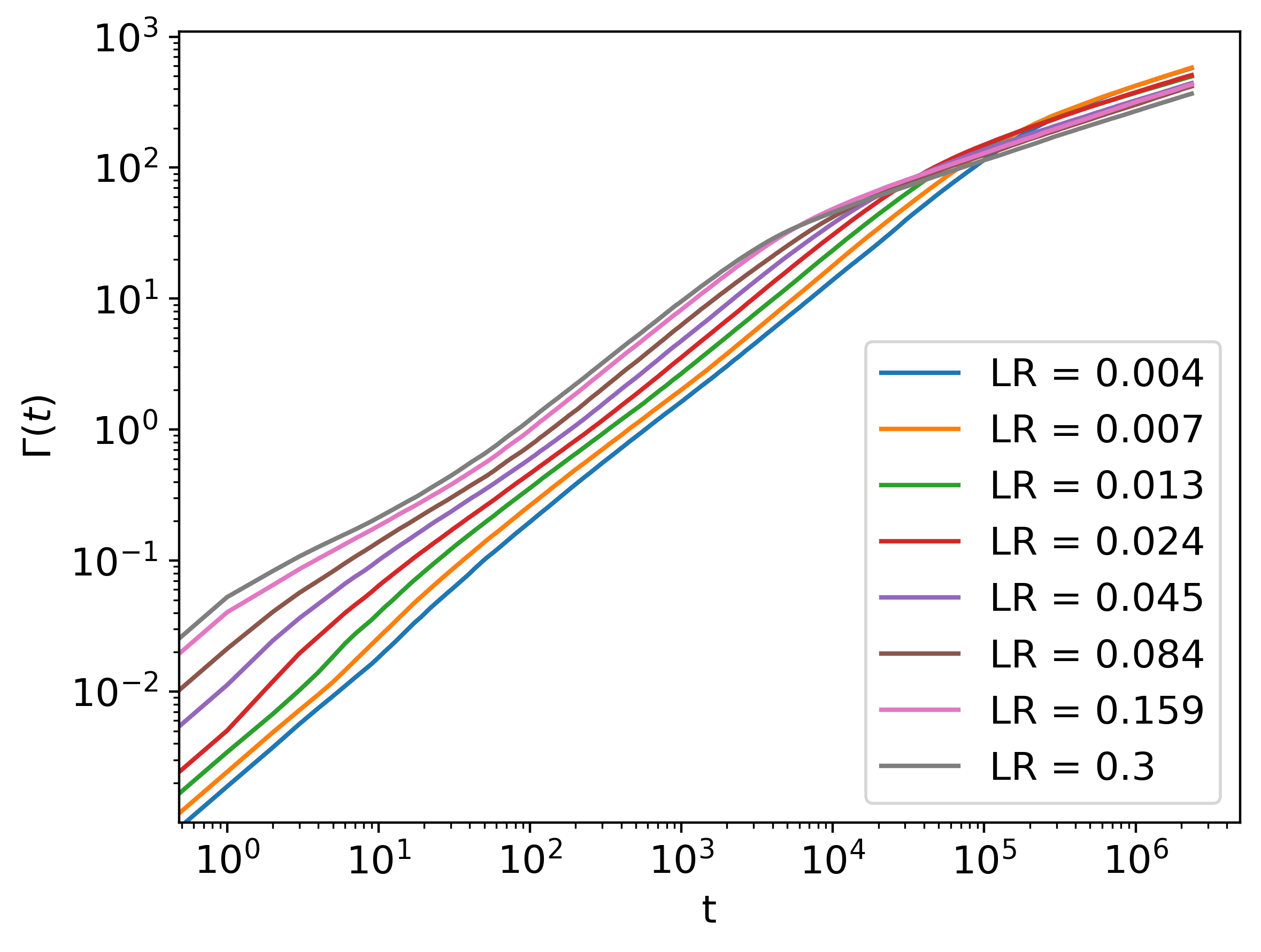}
  \end{subfigure}
  \caption{Information length for SGD \textbf{(left)} and Adadelta \textbf{(right)} optimizers with different learning rates on MNIST dataset.}
  \label{fig:mnist_lr_il_full}
\end{figure}

\begin{figure}[h]
  \begin{subfigure}[b]{0.49\textwidth}
    \includegraphics[width=\textwidth]{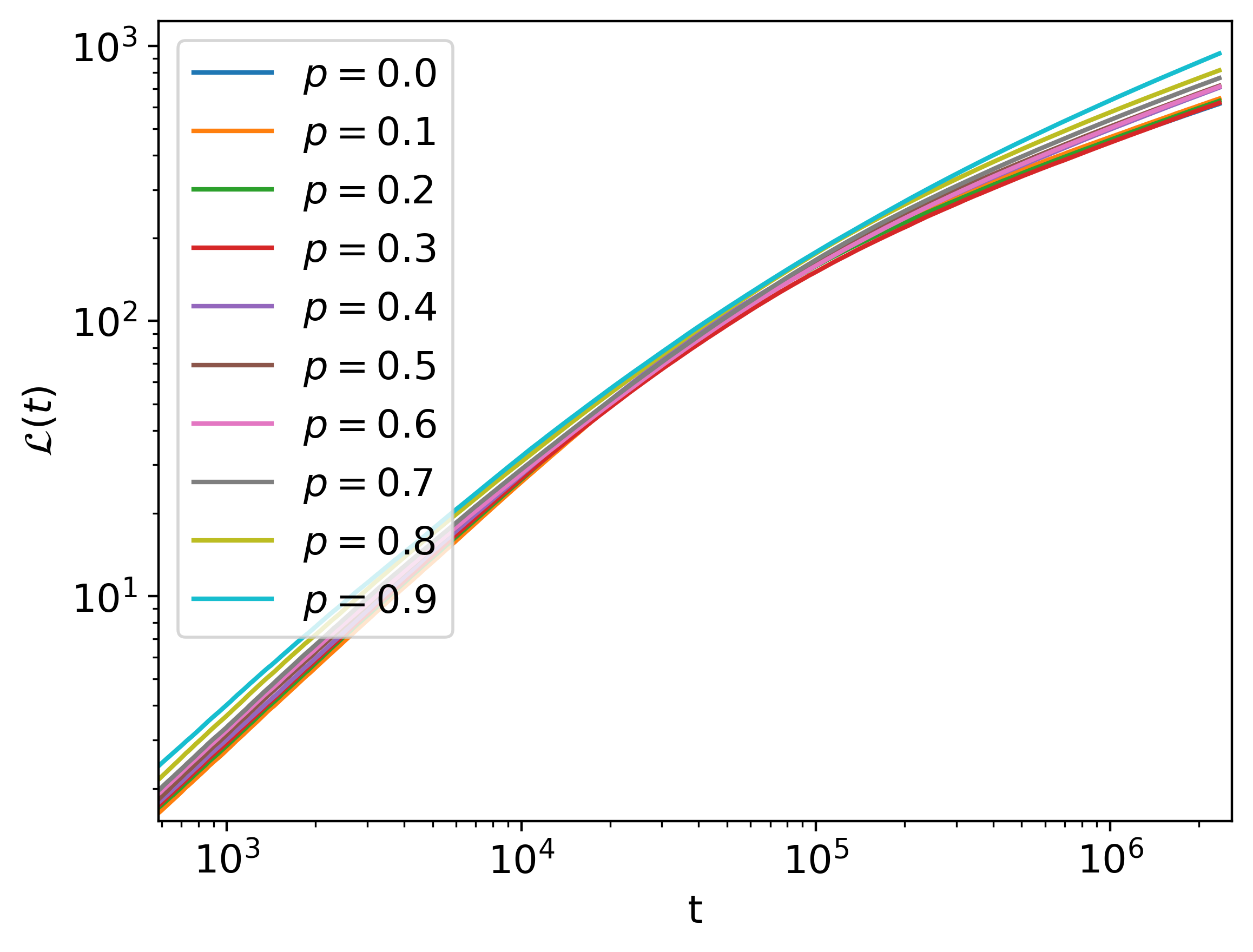}
  \end{subfigure}
  \hfill
  \begin{subfigure}[b]{0.49\textwidth}
    \includegraphics[width=\textwidth]{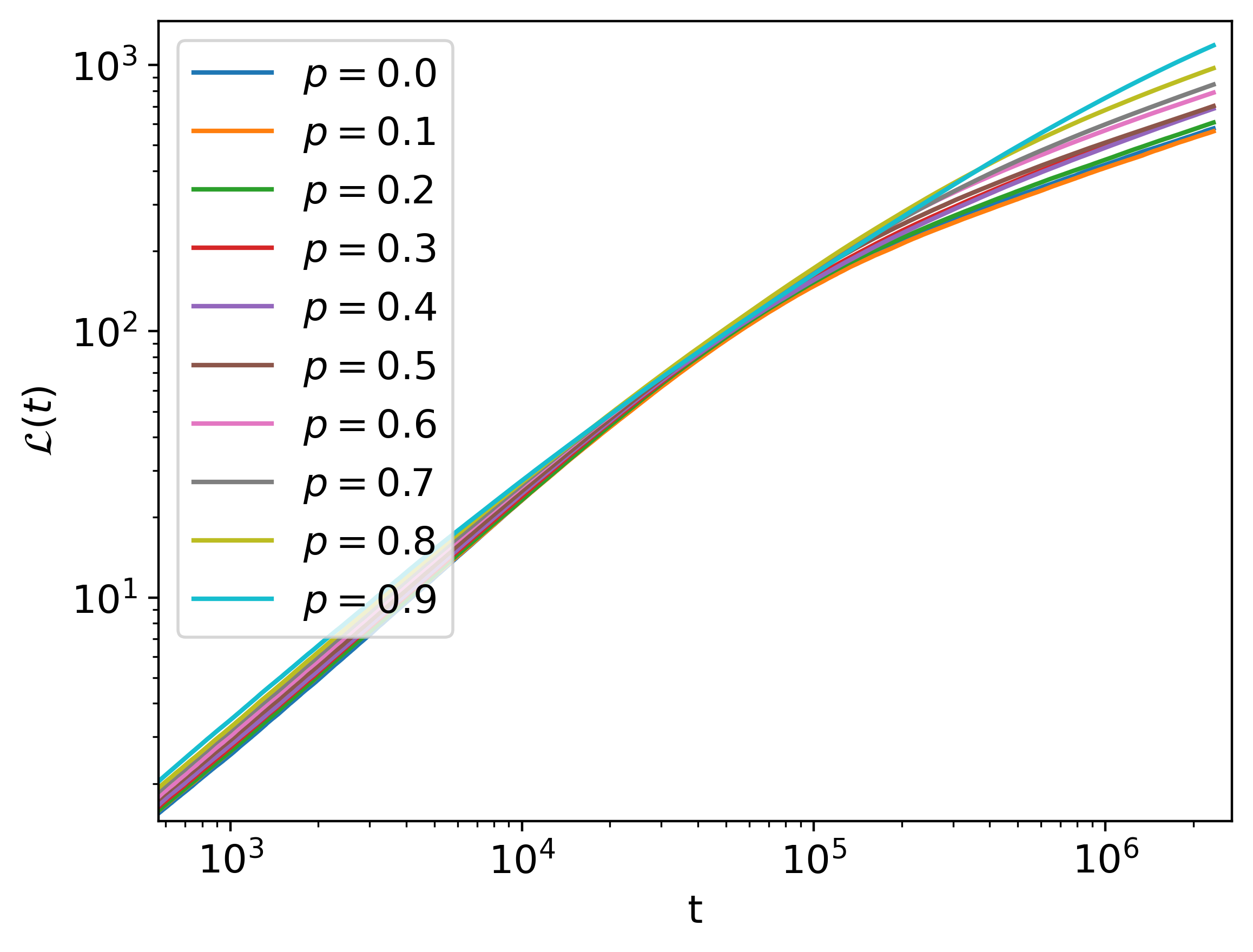}
  \end{subfigure}
  \caption{Information length for SGD \textbf{(left)} and Adadelta \textbf{(right)} optimizers with different dropout probabilites $p$ on MNIST dataset.}
  \label{fig:mnist_dropout_il}
\end{figure}

{
\begin{figure}[h]
    \centering
    \includegraphics[width=0.49 \textwidth]{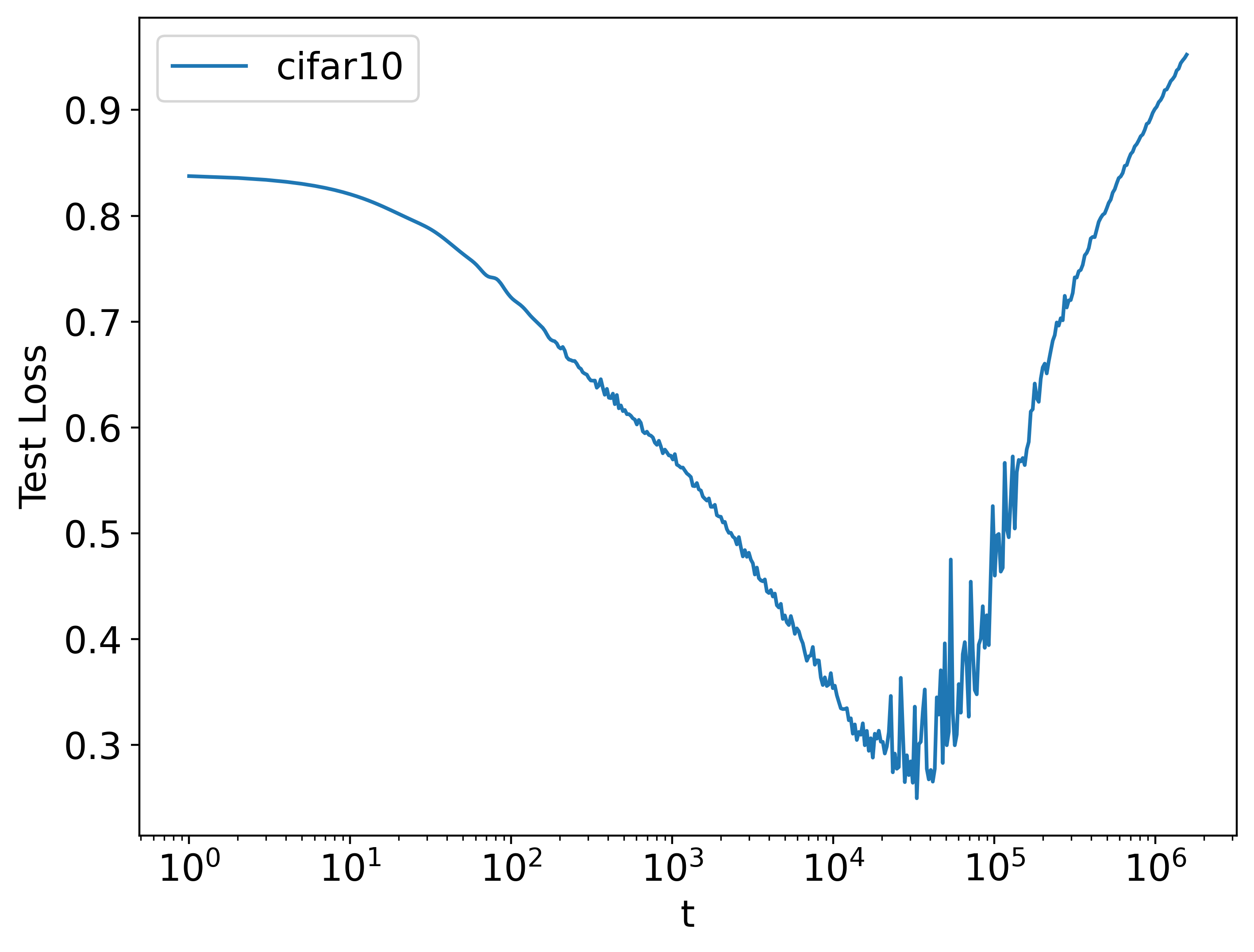}
  \caption{Test loss for ResNet-50 trained on CIFAR-10 dataset using SGD optimizer with learning rate 0.013.}
  \label{fig:resnet_test_loss}
\end{figure}
}

\begin{figure}[h]
  \begin{subfigure}[b]{0.49\textwidth}
    \includegraphics[width=\textwidth]{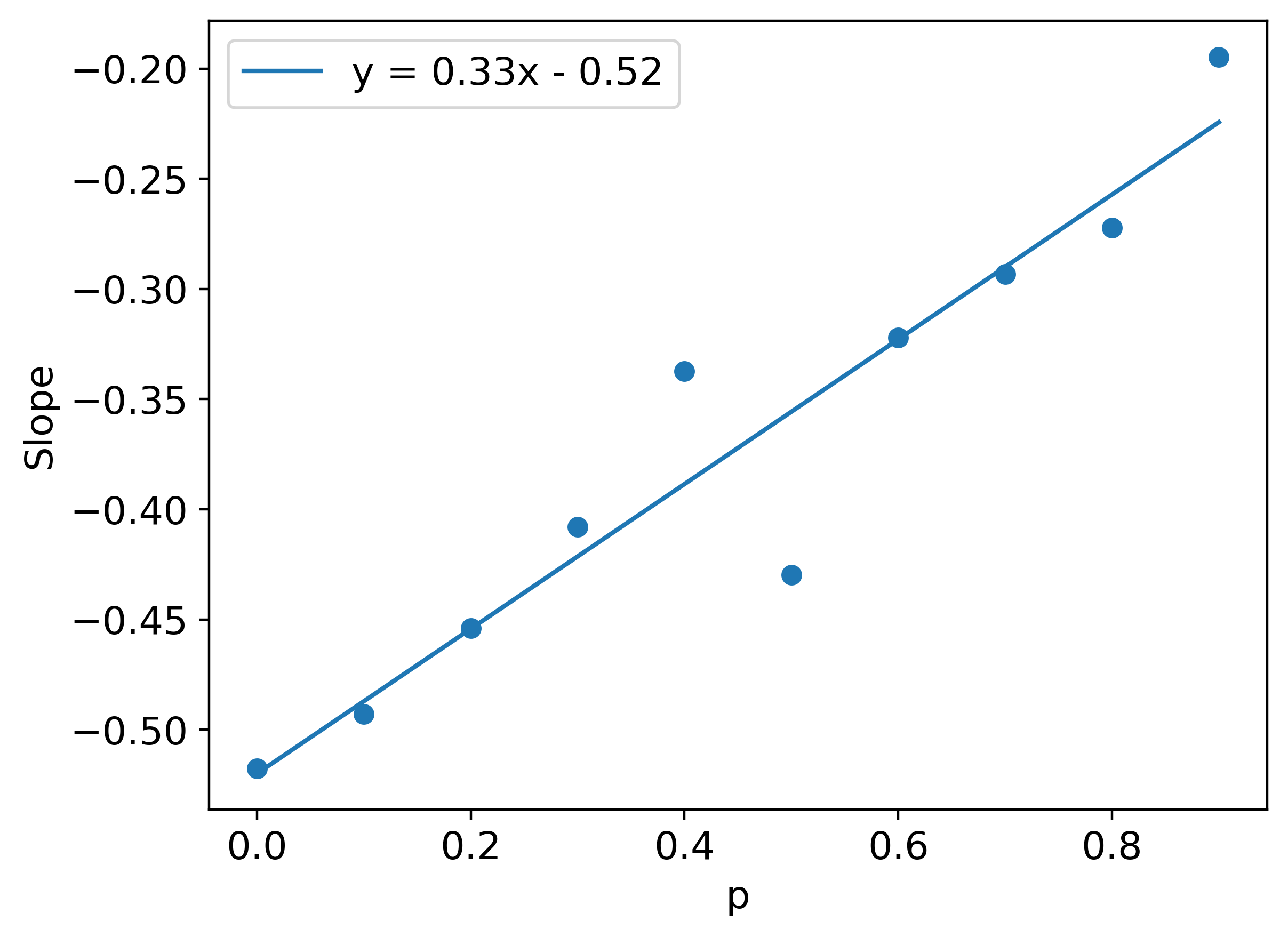}
  \end{subfigure}
  \hfill
  \begin{subfigure}[b]{0.49\textwidth}
    \includegraphics[width=\textwidth]{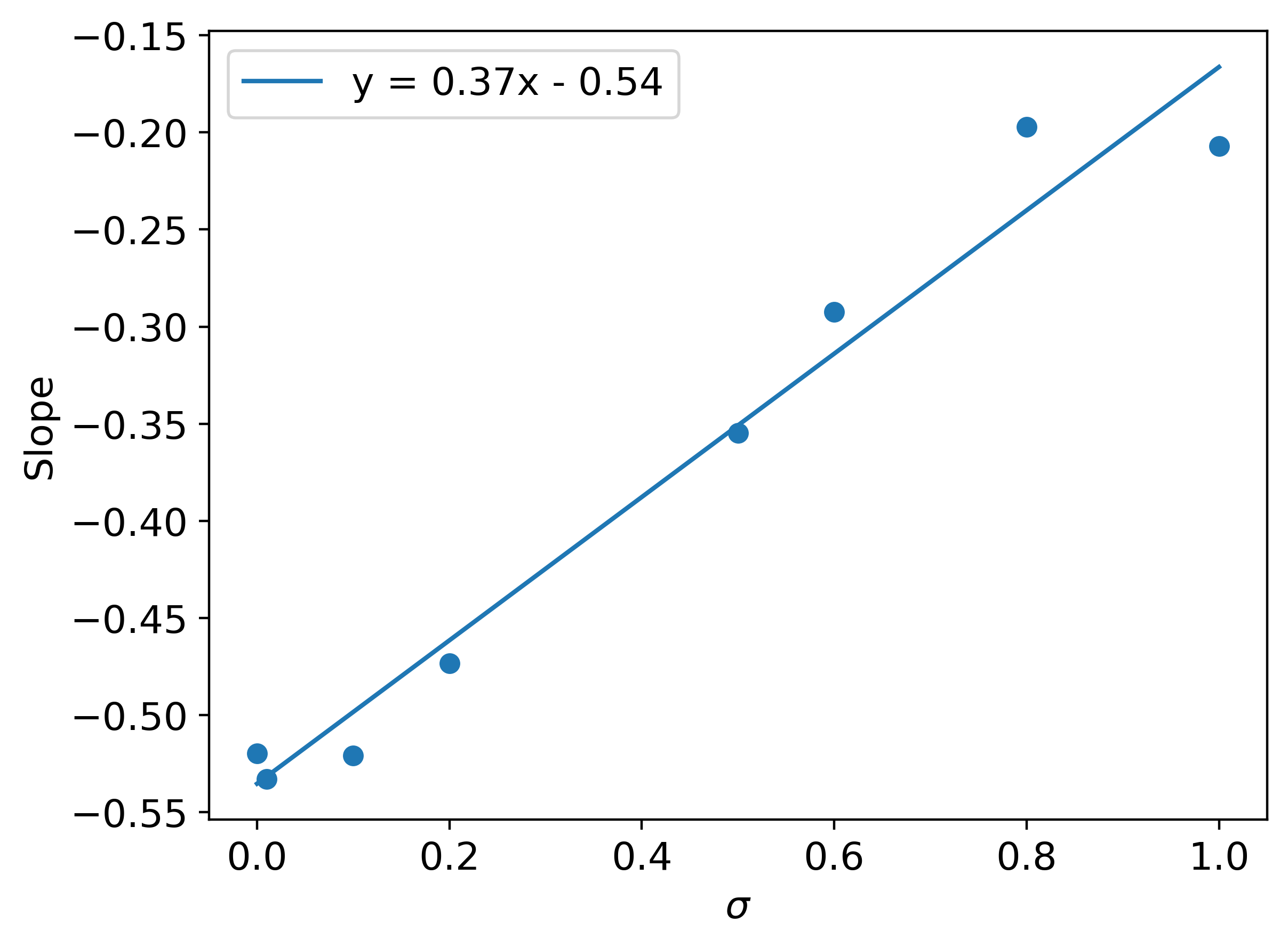}
  \end{subfigure}
  \caption{Slopes of the plot $d(\log \mathcal{L}) / d(\log t)$ vs. $\log t$ (Fig. \ref{fig:adadelta_dropout_std_deriv1}) at $t$ where test loss is minimum, for Adaldelta with lr = 0.013 with different \textbf{(left)} dropout and \textbf{(right)} noise levels trained on the MNIST dataset. The slopes were computed by taking 40 data points around the minimum and performing a linear fit.}
  \label{fig:adadelta_dropout_std_scaling}
\end{figure}

\FloatBarrier

\bibliography{main}

\end{document}